\documentclass[a4paper,fleqn]{cas-sc}

\usepackage[authoryear]{natbib}
\usepackage{subfigure}
\usepackage{amsthm}
\usepackage[section]{placeins} 
\def\tsc#1{\csdef{#1}{\textsc{\lowercase{#1}}\xspace}}
\tsc{WGM}
\tsc{QE}
\tsc{EP}
\tsc{PMS}
\tsc{BEC}
\tsc{DE}

\newtheorem{prop}[]{Proposition}

\newtheorem{lm}[]{Lemma}
\hypersetup{
 colorlinks,
 citecolor=blue,
 linkcolor=blue,
 urlcolor=blue}
\begin{document}
\let\WriteBookmarks\relax
\def\floatpagepagefraction{1}
\def\textpagefraction{.001}

\shorttitle{On the Robotic Uncertainty of Fully Autonomous Traffic}

\shortauthors{Li \textit{et al.}}

\title [mode = title]{On the Robotic Uncertainty of Fully Autonomous Traffic:\\ From Stochastic Car-Following to Mobility–Safety Trade-Offs}      



%
\author[1,2]{Hangyu Li}[orcid=0000-0002-8667-9928]


\author[1]{Xiaotong Sun}[orcid=0000-0002-3493-8828]

\cormark[1]

\author[1]{Chenglin Zhuang}[]

\author[2]{Xiaopeng Li}[orcid=0000-0002-5264-3775]

\affiliation[1]{organization={Intelligent Transportation Thrust, Systems Hub, 
 The Hong Kong University of Science and Technology (Guangzhou)},
 addressline={Nansha}, 
 city={Guangzhou},
 postcode={511458}, 
 state={Guangdong},
 country={China}}

\affiliation[2]{organization={Department of Civil \& Environmental Engineering, University of Wisconsin-Madison},
 city={Madison},
 postcode={53706}, 
 state={Wisconsin},
 country={United States}}

\cortext[cor1]{Corresponding author. Email:\textit{xtsun@ust.hk}}



\begin{abstract}
Recent transportation research highlights the potential of autonomous vehicles (AV) to improve traffic flow mobility as they are able to maintain smaller car-following distances. However, as a unique class of ground robots, AVs are susceptible to robotic errors, particularly in their perception and control modules with imperfect sensors and actuators, leading to uncertainties in their movements and an increased risk of collisions. Consequently, conservative operational strategies, such as larger headway and slower speeds, are implemented to prioritize safety over mobility in real-world operations. To reconcile the inconsistency, this paper presents an analytical model framework that delineates the endogenous reciprocity between traffic safety and mobility that arises from AVs' robotic uncertainties. Using both realistic car-following data and a stochastic intelligent driving model (IDM), the stochastic car-following distance is derived as a key parameter, enabling analysis of single-lane capacity and collision probability. A semi-Markov process is then employed to model the dynamics of the lane capacity, and the resulting collision-inclusive capacity, representing expected lane capacity under stationary conditions, serves as the primary performance metric for fully autonomous traffic. The analytical results are further utilized to investigate the impacts of critical parameters in AV and roadway designs on traffic performance, as well as the properties of optimal speed and headway under mobility-targeted or safety-dominated management objectives. Extensions to scenarios involving multiple non-independent collisions or multi-lane traffic scenarios are also discussed, which demonstrates the robustness of the theoretical results and their practical applications.
\end{abstract}



\maketitle

\section{Introduction}\label{sec:1}
The emergence of autonomous vehicles (AVs), or fully automated vehicles, offers the potential to improve the overall safety and mobility of road transportation \citep{fernandes2012platooning, jimenez2016advanced}. AVs with intelligent decision-making abilities and accurate machinery operations will be able to prevent collisions caused by human misbehavior, which is identified as the leading cause of car crashes \citep{mueller2020humanlike}. When investigating their impact on traffic mobility, AVs are usually treated as ideal machines with more advanced driving capabilities. Accordingly, they are assumed to maintain a shorter stable headway in traffic streams than human-driven vehicles (HDVs) \citep{morando2018studying}, contributing to different levels of increased roadway capacity \citep{ran1996traffic} as per their market penetration \citep{talebpour2016influence, chen2017towards, seo2017endogenous}
\footnote{\citet{seo2017endogenous} has shown AVs could improve the roadway capacity when their proportion reaches beyond some threshold.}.

However, AVs' robotic nature may prevent these benefits from being realized \citep{li2022equilibrium}. Like other mobile robots, AV operation is streamlined into a standard process that consists of four modules: perception, localization, planning (or decision-making), and control \citep{levinson2011towards}. While the first two modules rely on sensors that determine how accurately the AVs perceive their relative positional relationship with their surroundings, the latter two represent their level of driving intelligence. Each module contains a certain level of uncertainty, which together cause a chance of system error, leading an AV a deviation from its ideal movement or even a collision with the surrounding objects. We refer to these inherent errors as \textit{robotic uncertainties}, as they are problems that all autonomous robot systems will encounter, but have nothing to do with unknown external environments or intentions. Adopting conservative driving strategies, such as slower speed and longer car-following distance, could alleviate the negative impact of AVs' systematic uncertainty on driving safety performance \citep{sybis2020influence}. As a trade-off, it would limit the traffic capacity and undermine the overall traffic mobility.

Research efforts have been made to overcome the vehicular robotic deficiency, by reducing perception errors, improving localization accuracy \citep{wen2022tm3loc}, and developing robust control technologies against uncertainty \citep{huang2024human}, especially in Adaptive Cruise Control (ACC) or Cooperative Adaptive Cruise Control (CACC) \citep{zhou2017rolling,zhou2019robust,kontar2022bayesian}. Nevertheless, the system uncertainty could only be mitigated instead of being completely diminished \citep{mohammadian2023continuum}. In this respect, a growing research direction focuses on the verification and validation of AV safety performance under both simulated and realistic naturalistic driving environments \citep{waymo2017waymo, general2018self,feng2021intelligent,yan2023learning,xu2023safebench,feng2023dense,araujo2023testing}, aiming to identify the critical scenarios that lead to AV safety hazard. Also closely intertwined with uncertainty, stability emerges as another key concern. From the empirical experiments of ACC systems in commercial vehicles, \citet{makridis2021openacc} and \citet{li2021car} identified the characteristics of oscillation amplification. \citet{qin2023stability} further combine AV with HDV and study the stability and management of platoons under mixed traffic conditions. Other studies, such as \citet{zhang2020effect} and \citet{wang2020stability}, theoretically delineated the impact of uncertainty, particularly the positioning errors caused by robotic characteristics, on stability in automated car following systems. Traffic mobility, unfortunately, is usually overlooked in these studies. In fact, due to the absence of laws and regulations on AV traffic mobility \citep{shladover2019regulatory,vignon2025safety} and public concerns about autonomous driving collisions \citep{kyriakidis2015public,howard2014public,xing2022bicyclists}, autonomous driving companies usually adopt conservative driving approaches in real-road AV driving tests to achieve error-free safety performance. And as indicated earlier, when driving slower and keeping longer car-following distances than the surroundings, those pilot AVs naturally become moving bottlenecks that hold up regular traffic flows, generating traffic congestion and potential danger to roadways \citep{knoop2019platoon,schakel2017driving, mccarthy2022autonomous}.

This paper therefore aims to investigate the mutual relationship between traffic mobility and safety performance in a fully autonomous vehicle environment, e.g., a dedicated lane on a highway. In the literature, two recent studies have also highlighted the trade-off between AV safety and mobility under uncertainty \citep{shi2021empirical,li2022trade}. \citet{shi2021empirical} validate the presence of stochastic behavior of autonomous vehicles from empirical vehicle trajectory data under different headway settings. \citet{li2022trade} further delineate the trade-off through an analytical approach grounded in a mathematical model of vehicle control dynamics,  which also captures the relationship between stability and the two focal aspects: safety and mobility. This study further contributes from two perspectives. First, an analytical model is developed to explicitly incorporate the common robotic uncertainties arising from AV operation process, providing a structured approach to mathematically relate safety and mobility performances, which can be applied to both microscopic control models and empirical data. Second, macroscopic traffic flow considerations are emphasized beyond microscopic vehicle controls, connecting individual AV behavior to roadway-level outcomes, seeking to provide insightful suggestions to AV manufacturers and traffic management authorities to ensure AV development aligns with societal benefits.

The rest of this paper is organized as follows. Section~\ref{sec:2} overviews the model framework we proposed to link robotic uncertainties of autonomous vehicles with mobility of fully autonomous traffic, that is, the traffic streams composed by homogeneous autonomous vehicles. Under this framework, Section~\ref{sec:3} concentrates on the automated car-following process as a specific example and derives the stochastic motion in a statistic manner. Section~\ref{sec:4} then outlines the stochastic evolution of macroscopic fully autonomous traffic through a semi-Markov process and formulates the collision-inclusive capacity as an indicator for mobility. Based on the previous analytical results, Section~\ref{sec:5} discusses the applications of the model, including sensitivity analyses from the angle of design suggestions, and the optimization of critical variables in the view of transportation management. Section~\ref{sec:6} extends the problem to some special cases with extremely low probabilities to highlight the comprehensiveness and scalability of this study. Section~\ref{sec:7} then finally concludes the paper.

\section{Robotic Uncertainty and Collision-Inclusive Capacity} \label{sec:2}
Perception error is regarded as the vital error source that significantly impacts the motion of autonomous vehicles \citep{liu2021seeing}. Perception information is usually modeled as random variables with specific distributions, inducing to a probability of deviation between observation and reality. When the deviated observation is fed into the subsequent modules, the error propagates through the AV operation process, resulting in a stochastic deviation between the actual motion and the expectation. In addition, the controller is physically limited and cannot achieve the planned motion perfectly and without delay. Moreover, due to the complex vehicle dynamics and the large amount of randomness present in real-world operations, actuators can also generate certain errors when being controlled by autonomous driving \citep{zhou2017rolling,hu2022distributed}. These robotic uncertainties are inherent in all mobile robots with sensors, onboard algorithms, and actuators, even in an environment within operational design domain following established intentions. Therefore, even employing a theoretically collision-free driving strategy, AVs with robotic uncertainties may still have a chance to encounter collisions or any other unexpected behaviors. This characteristic and its impact are often ignored in the macroscopic traffic analysis of autonomous vehicles.

The following set of equations conceptually models the process described above. As AVs' operation is conducted in a discrete dynamic process, the ego vehicle $e$ estimates its surrounding objects' location in each time step $k$. Then the stochastic relative position $X^i_k$ of each object $i$ is observed based on the actual relative position ${\chi}^i_k$ with a random error $O({\chi}^i_k)$.
\begin{subequations}
 \begin{flalign}
  & \quad X^i_k = {\chi}^i_k + O({\chi}^i_k) ,\quad \forall i=1,2,...,n. \label{eq:percep_gen}
 \end{flalign}
 
The ego vehicle's planning could be integrated as a function $\Gamma(\cdot)$, which maps the observed historical and current relative positions of surrounding object $i,\: i=1,2,...n$ being expressed by $X^{1:n}_{k-m_h:k}$ to its planning trajectory $F_{k:k+m_f}^e$:
\begin{flalign}
 & F_{k:k+m_f}^e = \Gamma(X^{1:n}_{k-m_h:k}). \label{eq:plan_gen} 
\end{flalign}

The observed positions from $m_h$ steps ago ($k-m_h$) to the current step $k$ are utilized to optimize the perception results, make predictions of surrounding objects, and ultimately influence decision-making result. The planning trajectory $F_{k:k+m_f}^e$ looks $m_f$ steps ahead so as to generate a motion trajectory from current step $k$ to future ($k+m_f$). Note that the function itself is deterministic once the input is given, but the actual action $A_e^k$ is formed by the motion planning at the current time step $F_k^e$, combined with the control error $O(F_k^e)$:
\begin{flalign}
 & A^e_k = F_k^e + O(F_k^e). \label{eq:control_gen} 
\end{flalign}

As a random variable representing the position of the ego vehicle at time step $k+1$, $X^e_{k+1}$ is achieved by adding the ego vehicle's actual position ${\chi}^e_k$ with its stochastic motion $A^e_k$ at time step $k$:
\begin{flalign}
 & X^e_{k+1} = {\chi}^e_k + A^e_k. \label{eq:pos_gen} 
\end{flalign}
\end{subequations}

We denote $p^e_{k+1}$ as the probability of collision caused by the ego vehicle's stochastic motion at time step $k+1$, which is obtained by integrating the probability density $f$ of the stochastic position $X^e_{k+1}$ of the ego vehicle at that time step over the physical region $\Omega$, where the ego vehicle would collide with other traffic participants:
\begin{subequations}
\begin{flalign}
 & p^e_{k+1} = \int\limits_{\Omega} f_{X^e_{k+1}} (\omega) d\omega. \label{eq:p_gen} 
\end{flalign}

Due to the closed-loop control adopted by autonomous vehicles, their stability over time and over string can be strategically ensured. This property was proved in several previous studies \citep{gunter2020commercially,wang2020stability,li2022trade}. Accordingly, in the traffic stream with homogeneous autonomous vehicles, the collision probability of different vehicles at different time steps has a fixed expected value:
\begin{flalign}
 & \mathbb{E}p^e_{k+1} = p. \label{eq:avgp_gen} 
\end{flalign}

The macroscopic safety level $P$ can then be evaluated by comprehensively considering the collision probability of autonomous vehicles per unit range and over unit time $H$ in the transportation system. This depends on the vehicle density $\rho$ in the system, the time step of autonomous vehicle operations $\tau$, and the collision probability of a single vehicle in a single time step $p$:
\begin{flalign}
 & P = \rho\frac{H}{\tau}p. \label{eq:P_gen} 
\end{flalign}
\end{subequations}

Note that $H$ denotes the duration over which the transportation system is being analyzed. Since we focus on homogeneous autonomous vehicle streams with stability strategically ensured, the per-time-step collision probability of a single vehicle, $p$, is assumed to be independent of the choice of $H$.

Traffic mobility, on the other hand, is typically measured by traffic throughput, representing the number of vehicles passing through a designated study area, commonly a roadway segment, within a given time period. Traffic throughput is influenced by both roadway capacity and travel demand. In this study, however, we focus solely on roadway capacity or maximum throughput, as the demand is exogenously provided by social activities rather than being influenced by vehicle movements. When accounting for the possibility of collisions, the traffic state is either in normal operations with maximum capacity or in abnormal states when collisions block the traffic making throughput. The maximum capacity at normal state $s^{+}$ is determined by the driving policy, i.e., the decision-making process $\Gamma(\cdot)$ shown in Eq~\eqref{eq:plan_gen}, while the abnormal capacity $s^{-} = 0$. As a result, capacity at each fixed location becomes a random variable, whose expectation can be calculated based on the distribution over normal and abnormal states. We refer to this expectation as \textit{collision-inclusive capacity} (CIC), which can be mathematically provided as follows:
\begin{subequations}\label{eq:s_gen}
\begin{flalign}
 & s = (1-\lambda) s^{+} + \lambda s^{-},\\
 &\lambda=\Lambda(P).
\end{flalign}
\end{subequations}

Here, $\lambda$ represents the probability of being in the abnormal state. One should note that a location can be in an abnormal state either when a collision occurs directly at that location, or when collisions happen downstream that congest the traffic, or when collisions occur upstream and obstruct traffic moving downwards. In this regard, we use function $\lambda=\Lambda (P)$ to represent the relationship between the probability of being in the abnormal state and its influential factors. Clearly, $\frac{\partial \Lambda}{\partial P} \geq 0$ as a higher collision probability increases the chance of being in the abnormal state. Furthermore, $\Lambda(P=0)=0$, and $\Lambda(P\geq 1)=1$.

Besides, both capacity of the normal state $s^{+}$ and abnormal state probability $\lambda$ are affected by driving strategy $\Gamma(\cdot)$, as higher capacity indicates a shorter time gap between adjacent vehicles, which, according to Eq~\eqref{eq:P_gen}, also leads to a higher probability of collision. CIC then serves as a comprehensive measure of mobility in fully autonomous traffic.

\section{The Microscopic Car-Following Scenario}\label{sec:3}
We focus on the car-following scenario to further demonstrate the quantitative trade-off in mobility and safety. In this way, the single-lane CIC can represent mobility on the macroscopic scale. On the microscopic scale, we summarize the patterns of AV car-following with robotic uncertainties through simulation experiments and real AV data from Waymo \citep{Ettinger_2021_ICCV}, and then derive a random distribution representation to assist in subsequent analytical analysis.

\subsection{Scenario establishment}
Car-following is the simplest and most commonly seen traffic scenario, as well as the earliest and most mature vehicle automation functions. Several key assumptions on AVs' car-following scenarios are presented as follows:
\begin{figure}[!ht]
 \centering
 \includegraphics[width=0.85\textwidth]{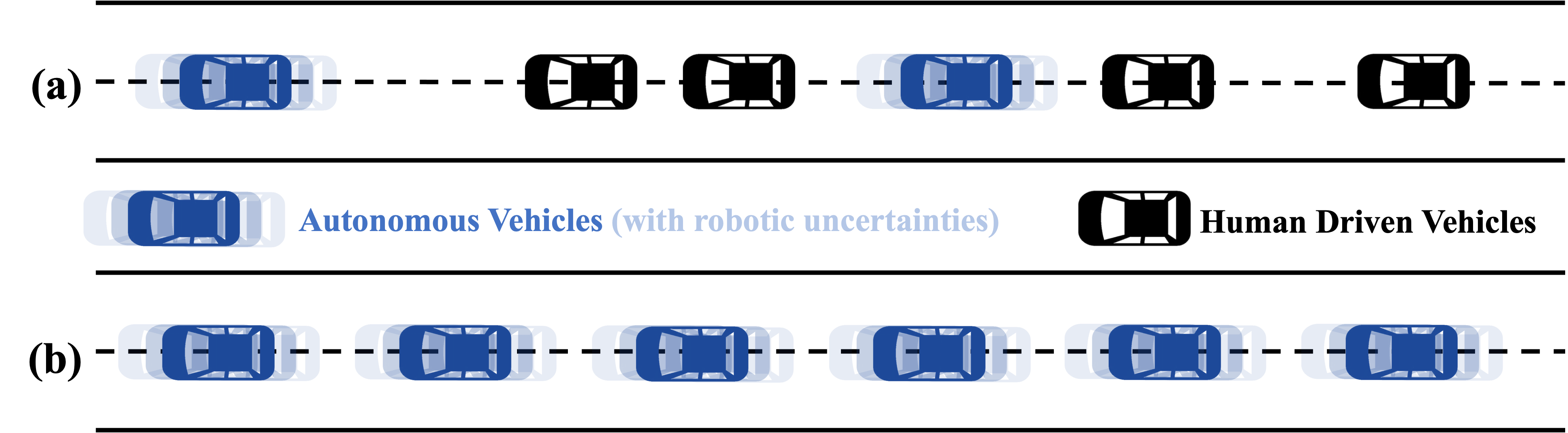}
 \caption{(a) Mixed traffic flow with both AVs and HDVs; (b) Fully autonomous traffic with robotic uncertainties.}
 \label{fig:car-following}
\end{figure}

\begin{enumerate}
 \item \textit{Ideal roadway segment.} Consider a basic road segment with neither on- and off-ramps nor increase and decrease in the lane number. An infinite number of AVs are assumed to be discharged from upstream, and there are no bottlenecks downstream of the segment. Therefore, when considering boundary conditions, the inlet boundary follows the flow conservation and the outlet boundary has no constraints. In addition, no cyclic impact is considered, though they may exist on some particular roads, such as roundabouts, where the situation is complex and beyond the scope of this paper.
 
 \item \textit{Longitudinal control.} Our study focuses on longitudinal car-following in which only acceleration and braking of the vehicle are considered, and lateral maneuvers are excluded. Furthermore, only the influence of the preceding vehicle is taken into account for the ego vehicle. Perceptual information other than measurements of the preceding vehicle is ignored. In the meantime, the assumed perfect lateral control does not introduce additional errors to longitudinal movements. With the previous assumptions, rear-end collisions are the only type of accident. The location on the lane where collisions happen will be directly blocked, reducing the throughput to zero. Furthermore, its influence would spread to both upstream (blocked) and downstream (empty) traffic.
 
 \item \textit{Predetermined Speed.} Human-driven vehicles usually adjust and maintain a safe and comfortable speed when their movements are restricted by vehicles in front. From the microscopic perspective, it leads to heterogeneous car-following distances, as shown in Figure~\ref{fig:car-following}(a). From an aggregate point of view, it contributes to the endogenous relationship of average speed and traffic density under equilibrium represented by the fundamental diagram \citep{greenshields1935study, daganzo1997fundamentals}. Alternatively, fully autonomous car-following can be programmed and predetermined. Although they have stochastic errors due to robotic uncertainties, typical or average values of different speeds and following distances can be maintained, as shown in Figure~\ref{fig:car-following}(b). Therefore, we view speed as an exogenous variable whose relationship with traffic density can be programmed.
\end{enumerate}

\subsection{Simulation of a high-order car-following model}
\begin{figure}[htpb]
 \centering
 \includegraphics[width=0.9\textwidth]{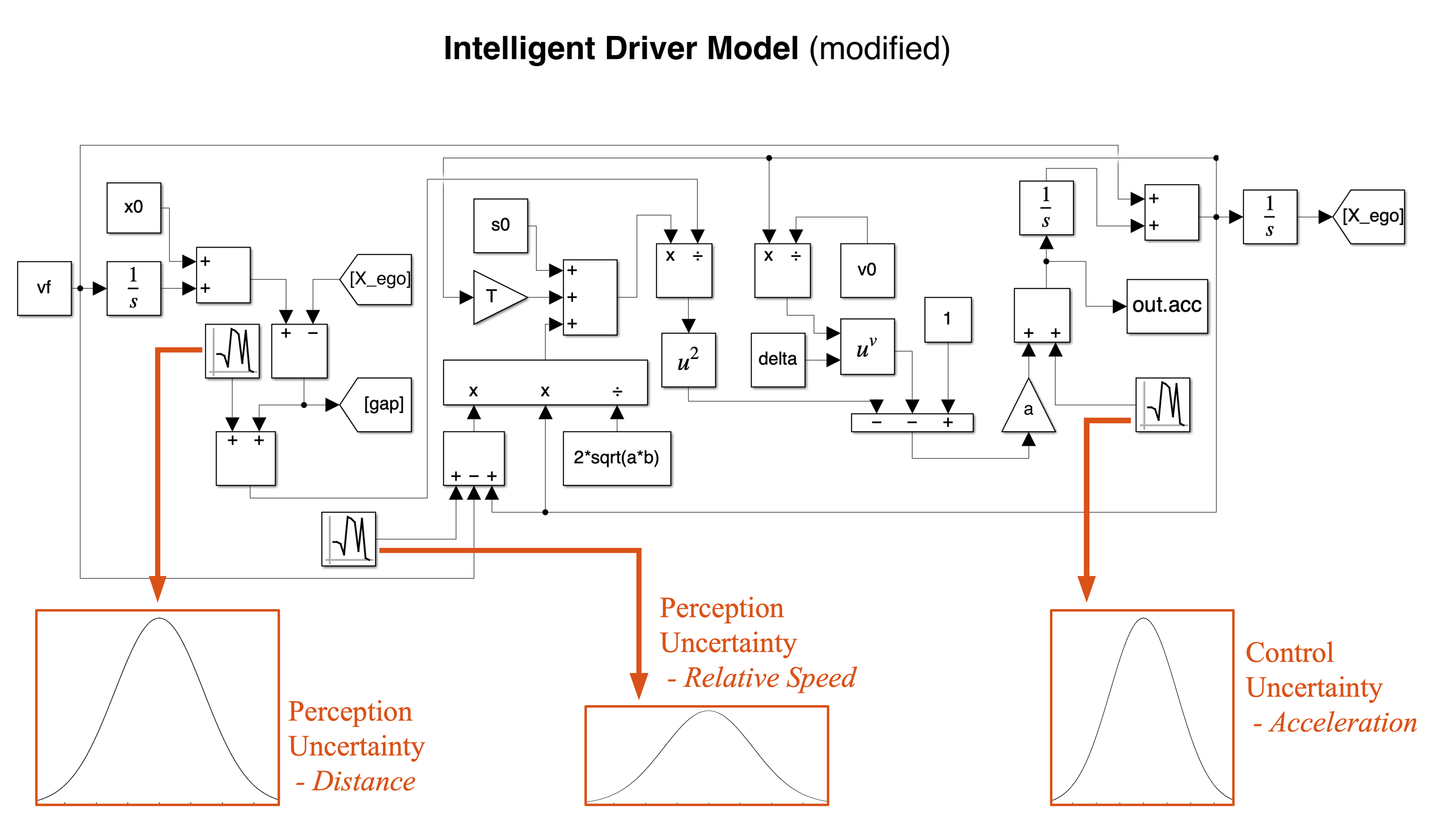}
 \caption{Simulated stochastic car-following motion under robotic uncertainties based on IDM car-following model. The dynamics are simulated by MATLAB Simulink, and robotic errors are introduced into the perception of distance, the perception of speed, control.}
 \label{fig:IDM}
\end{figure}

To quantify the impact of robotic uncertainty, we modify the Intelligent Driver Model (IDM), which is widely used in autonomous vehicle analysis and is believed the foundational model for various Adaptive Cruise Control (ACC) functions \citep{milanes2014modeling}, by adding independent stochastic terms to an ego vehicle's observations and control following a preceding vehicle. The modification provides:
\begin{subequations}
 \begin{flalign}
  &\dot{x}_e = v_e, \label{eq:IDM_v}\\
  &\dot{v}_e = a \left(1- \left(\frac{v_e}{v_0} \right)^{\xi}-\left(\frac{d^*(v_{e}, \Delta v^o)}{d^o}\right)^2 \right)+\epsilon_{\dot{v}}, \label{eq:IDM_a}\\
  &d^*(v_{e},\Delta v^o)=d_0+v_e h_0+\frac{v_e\Delta v^o}{2\sqrt{a b}}. \label{eq:IDM_d*}
 \end{flalign}\label{eq:IDM}

In Eq~\eqref{eq:IDM_v}, $x_e$ indicates the location of the ego (following) vehicle, whose increment is its actual speed $v_e$. The acceleration $\dot{v}_e$ is bounded by the maximum acceleration and comfortable deceleration $a$ and $b$ in Eqs~\eqref{eq:IDM_a} and \eqref{eq:IDM_d*}. In addition, $v_0$, $d_0$, and $h_0$ refer to the free flow speed, minimum safe distance (or named as effective vehicle length by \cite{treiber2013traffic}), and safe headway, respectively. Under constant time headway (CTH) car-following policies, $d_0$ is set to zero. It leads to two observations on gap and speed difference, which satisfy the following system of stochastic differential equations (SDE):
 \begin{flalign}
  &d^o = d + \epsilon_{d}, \label{eq:IDMod}\\
  &\Delta v^o = \Delta v + \epsilon_{\Delta v}, \label{eq:IDMov}
 \end{flalign}\label{eq:IDMo}
\end{subequations}
where $\epsilon_{d}$ is the stochastic term for observation of distance, and $\epsilon_{\Delta v}$ the uncertainty on speed difference. Moreover, $\epsilon_{\dot{v}}$ demonstrates the control error in Eq~\eqref{eq:IDM_a}. They are all assumed to follow Gaussian distributions with zero mean \citep{ni2009sensor}.

Since there are no closed-form solutions for the above SDEs, we resort to simulation methods to obtain an AV's random motion. The simulation process is given in Figure~\ref{fig:IDM}. To start with, we established a basic IDM model and added corresponding perception and control errors to the observation of distance, relative velocity, and the ego vehicle's acceleration. These errors are independent and follow Gaussian distributions. We then ran this modified IDM model at a given time step (consistent with the control time step of a real AV), with the preceding vehicle being set a constant driving speed. In the simulation experiments, the preceding vehicle drives at $v=50$ km/h, while the free flow speed is set at $v_0=120$ km/h; The other key parameters are set to $a=2$ $\mathrm{m \cdot s^{-2}}$, $b=2$ $\mathrm{m \cdot s^{-2}}$, $\xi = 4$, $d_0 = 0$ m, and $h_0 = 1.5$ s. In each round of simulation, the initial car-following distance and speed of the ego vehicle are at their expected values, which are $v_e(0) = v_f = 50$ km/h and $d(0) = \left(d_0+h_0 v_e(0)\right) \left(1-\left(v_e(0)/v_0\right)^\xi\right)^{-\frac{1}{2}} = 21.1546$ m.

\begin{figure}[htpb]
 \centering
 \includegraphics[width=\textwidth]{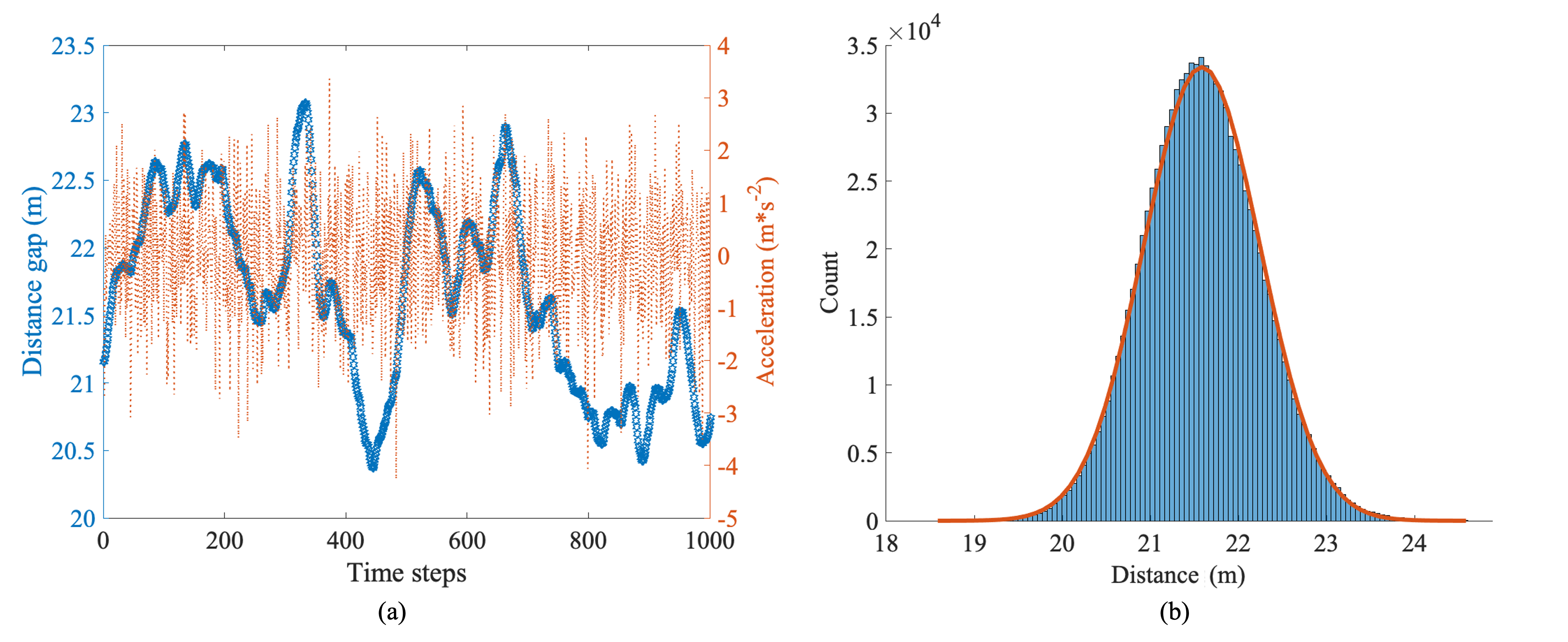}
 \caption{A typical simulation result: (a) Acceleration of the ego vehicle (red dotted line) and distance between two vehicles (blue circled line) over time; (b) Histogram of distance between two vehicles (blue bars) and the fitted Gaussian distribution (red line).}
 \label{fig:Result}
\end{figure}

Figure~\ref{fig:Result} illustrates the simulation results when all added error terms following the standard Gaussian distribution, that are $\sim \mathcal{N}(0,1)$. In particular, Figure~\ref{fig:Result}(a) shows the stochastic changes of ego vehicle's acceleration over time and its corresponding car-following distance. Figure~\ref{fig:Result}(b) then presents the histogram of distance between the two vehicles during the whole simulation process. As can be seen, the distance exhibits strong Gaussian characteristics in the statistical sense: A bell-shaped curve is formed where values closer to the mean have higher probabilities. The red line in Figure~\ref{fig:Result}(b) represents the probability density of a Gaussian distribution, fitted using the statistical mean and variance derived from the simulation results of the following distances. The visual comparison between the simulated results and the fitted Gaussian distribution suggests that the car-following distances under stochastic observation and speed exhibit Gaussian errors over time, despite the analytical form being unobtainable due to the closed-loop control characteristics of the IDM.

To further evaluate the goodness-of-fit of the Gaussian distribution in modeling stochastic car-following distance in a macroscopic statistical sense, Normalized Root Mean Squared Error (NRMSE) is used as the quantitative measure:
\begin{flalign}
 \text{NRMSE} = \frac{\|\overrightarrow{\textit{ref}}-\overrightarrow{\textit{exp}}\|}{\|\overrightarrow{\textit{ref}}-\textit{mean}(\overrightarrow{\textit{ref}})\|}. \label{eq:NRMSE} 
\end{flalign}

Here, $\| \cdots \|$ indicates the 2-norm of a vector, $\overrightarrow{\textit{ref}}$ refers to counts in each corresponding bin derived from the fitted Gaussian distribution, $\overrightarrow{\textit{exp}}$ shows the vector of counts in $100$ bins generated from the simulation experiment results, and $\textit{mean}(\cdot)$ calculates the average value of a vector. According to its definition, an NRMSE close to zero indicates a perfect fit to the reference data (i.e. no error), whereas an NRMSE close to one means a fit no better than a straight line at matching the reference.

To test the robustness of the fitted Gaussian under different uncertainty conditions, we assume that the observation and control errors $\epsilon_d$, $\epsilon_{\Delta v}$, and $\epsilon_{\dot{v}}$ have variances of $0.1$, $0.5$, and $1$, while their mean values remain at $0$. The typical values represent the universal applicability of the perception capability of AVs as they cover precision from upper limit (such as the state-of-the-art performance in the NuScenes perception benchmark; \citealp{nuscenes2019}) to lower limit (sub-meter level precision of millimeter wave radars). This results in a total of 27 fitness results, summarized in Table~\ref{tab:simulation}. The results indicate that NRMSEs are less than 0.06 regardless of the uncertainty conditions. Thus, it is reasonable to assume that the stochastic car-following distance follows a Gaussian distribution in a macroscopic statistical sense, and this assumption will be used for subsequent analysis.
\begin{table}[htbp]
 \caption{Goodness of fit of Gaussian distribution to distance under different uncertainty conditions.}
 \centering
 \begin{tabular*}{\tblwidth}{@{}@{\extracolsep{\fill}}l|ccc|ccc|ccc@{}}
  \toprule
  \multicolumn{1}{r|}{$\epsilon_{d} \sim$} & \multicolumn{3}{c|}{$\mathcal{N}(0,0.1)$}& \multicolumn{3}{c|}{$\mathcal{N}(0,0.5)$}& \multicolumn{3}{c}{$\mathcal{N}(0,1)$}\\
  \multicolumn{1}{r|}{$\epsilon_{\Delta v} \sim$} & $\mathcal{N}(0,0.1)$ & $\mathcal{N}(0,0.5)$ & $\mathcal{N}(0,1)$ & $\mathcal{N}(0,0.1)$ & $\mathcal{N}(0,0.5)$ & $\mathcal{N}(0,1)$ & $\mathcal{N}(0,0.1)$ & $\mathcal{N}(0,0.5)$ & $\mathcal{N}(0,1)$ \\
  \midrule
  $\epsilon_{\dot{v}} \sim \mathcal{N}(0,0.1)$ & 0.0170 & 0.0172 & 0.0208 & 0.0169 & 0.0168 & 0.0189 & 0.0166 & 0.0193 & 0.0198 \\
  $\epsilon_{\dot{v}} \sim \mathcal{N}(0,0.5)$ & 0.0410 & 0.0278 & 0.0226 & 0.0408 & 0.0280 & 0.0236 & 0.0392 & 0.0280 & 0.0237 \\
  $\epsilon_{\dot{v}} \sim \mathcal{N}(0,1)$ & \textbf{0.0590} & 0.0479 & 0.0408 & 0.0585 & 0.0478 & 0.0392 & 0.0579 & 0.0465 & 0.0388 \\
  \bottomrule
 \end{tabular*}
 \label{tab:simulation}
\end{table}

\subsection{Empirical AV data analysis}
Considering that simulation experiments may not fully capture realistic driving behaviors, we also utilized real-world car-following data from commercial automated vehicles to validate their stochastic characteristics in longitudinal movements and verify our modified IDM in describing AV stochastic motion caused by robotic uncertainties. The Waymo Open Dataset \citep{Ettinger_2021_ICCV} was selected as the primary dataset for the empirical analysis. \citet{zhou2024unified} processed the dataset, focusing on longitudinal car-following scenarios with a constant-speed leading vehicle (LV) to accurately reflect the intricate behaviors of AVs, distinguishing it from the stochasticity caused by LV's dynamics. Key data representing the states of the LV and the following AV (FAV), including spatial gaps, speeds, accelerations, and positions, are recorded. Our previous study \citep{zhou2024unified} identified that Waymo's data collection methods make it the only dataset capable of effectively capturing speed fluctuations in AV operations while maintaining stability. The OpenACC dataset \cite{makridis2021openacc} with multiple car-following AVs, however, exhibits string instability, and the use of smoothing postprocessing methods to makes it unable to capture subtle stochastic motions. Therefore, considering that we isolated the impact of robotic uncertainty from LV dynamics on FAV motion through data filtering, it is well-suited for this study in identifying reliable results of stochastic car-following distances.

To identify stable car-following scenarios from complex real-world AV driving data, we conducted a sequence of data filtering steps. First, based on our previous work \citep{zhou2024unified} extracting unified car-following data, we selected scenarios where the preceding vehicle of the FAV maintained a constant speed, while the FAV tried to keep a consistent distance. Specifically, we retain the cases where the LV maintains a steady speed of around $20.2m/s$, the most common speed observed for the LVs, with a deviation of less than $0.2m/s$. Next, we excluded cases where the speed difference between two vehicles exceeded $1$ m/s, as larger differences may indicate an unstable car-following state, such as a free-flow approach or continuous braking. Finally, since the default car-following distance setting could vary across driving scenarios, we normalized the car-following distance in all sampled data to have a uniform variance of one and a mean of zero, ensuring comparability across the sub-datasets.

\begin{figure}[htpb]
 \centering
 \includegraphics[width=\textwidth]{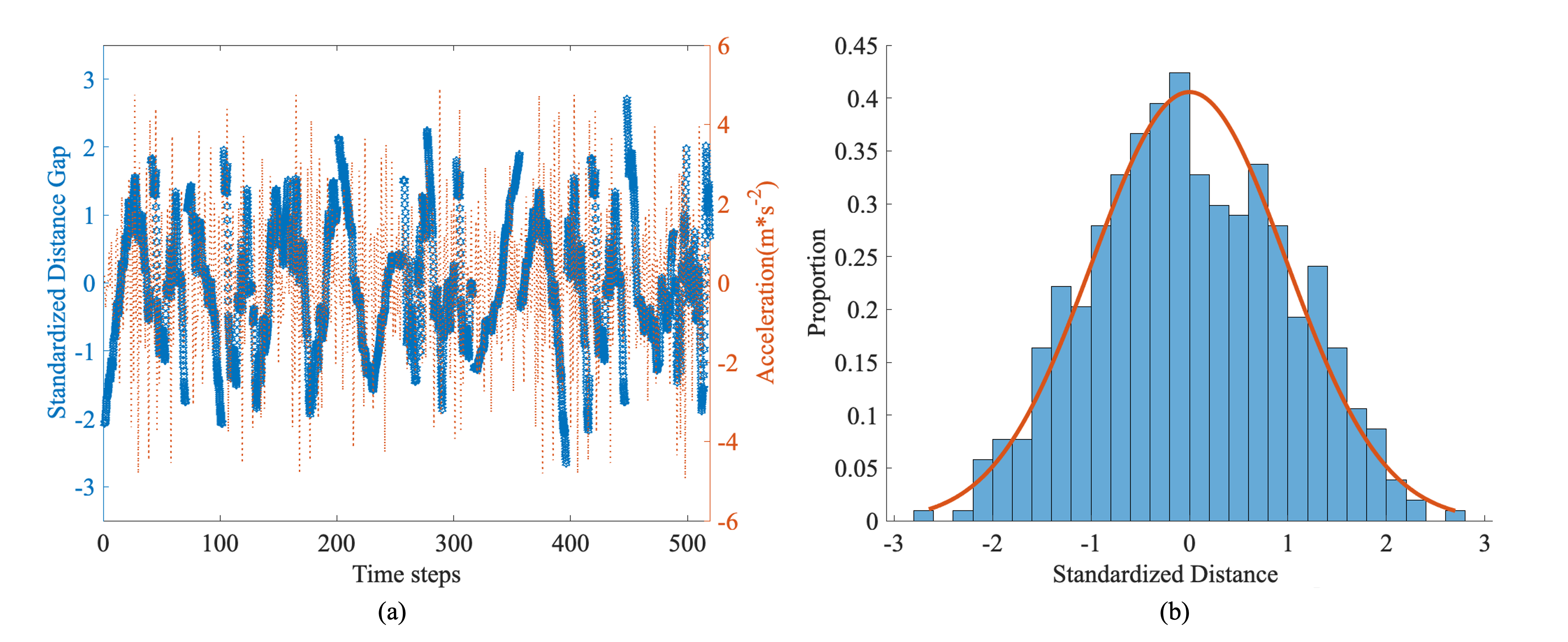}
 \caption{A real car-following data result from Waymo dataset: (a) Acceleration of FAV (red dotted line) and distance between two vehicles (blue circled line) over time; (b) Histogram of distance between two vehicles (blue bars) and the fitted Gaussian distribution (red line).}
 \label{fig:WaymoResult}
\end{figure}

Figure~\ref{fig:WaymoResult} illustrates the realistic car-following behaviors of AVs after data processing. In align with Figure~\ref{fig:Result},Figure~\ref{fig:WaymoResult}(a) displays the stochastic variations in the FAV’s acceleration over time, which is colored in red, along with the corresponding car-following distances of FAVs represented by blue circular dots. The observed discontinuities in the distance data are attributed to the inclusion of multiple trajectories. Figure~\ref{fig:WaymoResult}(b) presents a histogram of car-following distances across all sampled trajectories. As can be seen, the car-following distances in real driving scenarios also exhibit strong Gaussian characteristics in the macroscopic statistical sense: the histogram forms a bell-shaped curve where values closer to the mean have higher probabilities. Similarly, the red line illustrates the probability density of the fitted Gaussian distribution, using the mean and variance from the sampled data, with a NRMSE of 0.23348 quantifying the goodness-of-fit. Overall, the empirical data analysis suggests that Gaussian distribution can serve as a reasonable approximation of AVs' stochastic car-following distances, which validates the property revealed in the IDM simulation and also support its use in further analysis.

For other potential AV data in the future, or for those collected from new observations that may not strictly conform to the Gaussian distribution (generalized form as Eq~\eqref{eq:p_gen}), advanced system identification methods \citep{meng2024koopman} can help determine AV car-following behavior and distinguish between deterministic and stochastic components. Techniques such as Kernel density estimation can be adopted to derive an empirical probability density estimation. With the empirical estimation, say $\hat{f}(x)$, the collision probability in Eq~\eqref{eq:p_gen} can be derived by $\int_{-\infty}^{l}\hat{f}(x)dx$. Thanks to its analytical nature, in spite of the empirical density function, subsequent macroscopic analyses shall still be valid. It makes our whole analysis framework highly robust and universal, highlighting its significance for traffic design and management.

\subsection{Stochastic car-following distance in a stable string}
In this section, we extend the previous modified IDM model in Eqs~\eqref{eq:IDM} from an arbitrary pair of AVs to a homogeneous, fully autonomous car-following string, investigating the statistical properties of their car-following distances. By homogeneity, we mean that all vehicles adopt the same AV technology and driving strategy. The shared technology ensures a unified AV operation processing time $\tau$ and identical uncertainties $\epsilon_d$, $\epsilon_{\Delta v}$, and $\epsilon_{\dot{v}}$, while the same driving strategy guarantees that all vehicles maintain the same desired speed $v$ and headway $\eta$.

We assume that the car-following string remains stable when operating on dedicated lanes or within an operational design domain. This implies both the stability of each single vehicle's control and the overall string stability of the vehicle string \citep{feng2019string}. For detailed stability analysis of vehicle operations, we refer to our previous work \citep{wang2020stability,ma2022string,li2022trade}, which will not be elaborated here. Consequently, the variance of the stochastic car-following distance of each pair or vehicles does not amplify with the string index but instead converges to a value fixed value. Meanwhile, as both IDM simulations and empirical AV data analysis suggest a Gaussian distribution for stochastic car-following distances, we denote the the error between the desired and actual distance $\epsilon_x \sim \mathcal{N}(0,\sigma_x^2)$, where $x$ represents indicate any index in the string.
\begin{table}[htbp]
 \caption{R-Square values for different fitting to stochastic car-following distance.}
 \centering
 \begin{tabular}{cccccc|ccc}
  \toprule
  \multicolumn{6}{c|}{Polynomial}& \multicolumn{3}{c}{Exponential}\\
  \midrule
  $v \eta$ & $v \eta^2$ & $v \eta^3$ & $v^2 \eta$ & $v^2 \eta^2$ & $v^3 \eta$ & $e^{v \eta}$ & $e^{v \eta^2}$ & $e^{v^2 \eta}$ \\
  \midrule
  0.9288 & 0.9122 & 0.7463 & \textbf{0.932} & 0.9202 & 0.7703 & 0.6725 & 0.7592 & 0.7675 \\
  \bottomrule
 \end{tabular}
 \label{tab:distance_fitting}
\end{table}

\begin{figure}[htpb]
 \centering
 \includegraphics[width=0.65\textwidth]{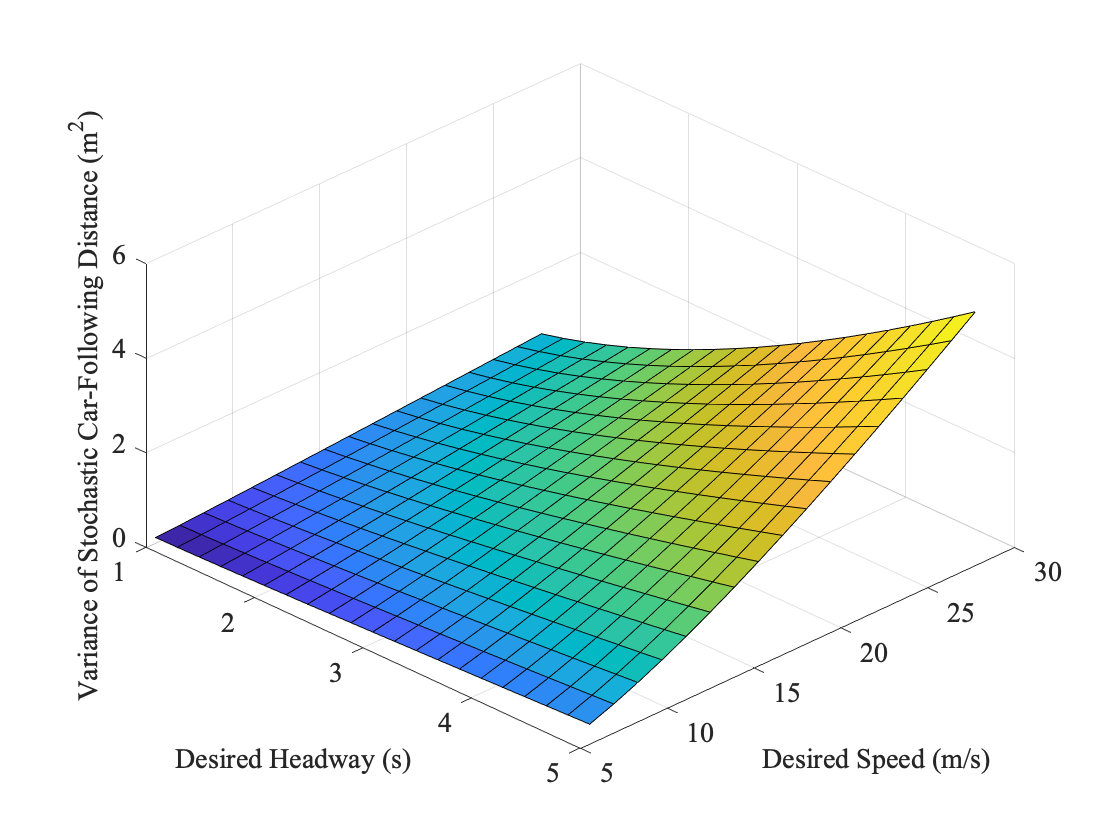}
 \caption{Relationship between the variance of stochastic car-following distance with desired speed and headway.}
 \label{fig:distance_var}
\end{figure}

We further explore the relationship between $\sigma_x^2$ and the two controllable variables, desired speed $v$ and headway $\eta$, using IDM simulations. Figure~\ref{fig:distance_var} visualizes how $\sigma_x^2$ changes as the desired speed varies from $v=20$ km/h to $100$ km/h and $\eta$ from $1.0$ s to $5.0$ s. Correspondingly, a set of regressions are conducted to determine the most suitable quantitative relationship. The R-Square values are shown in Table~\ref{tab:distance_fitting}, suggesting that $\sigma_x$ is most likely to be proportional to $v^2\eta$. In addition, we notice that the desired headway $\eta$ and speed $v$ can be independently controlled under fully autonomous traffic. In this context, the stochastic car-following distance can then by formulated as follows:
\begin{subequations} \label{eq:d_v_eta}
\begin{flalign}
 &d(v,\eta) = v\eta + \epsilon_x, \\
 &\epsilon_x \sim \mathcal{N}(0,\sigma_x^2), \\
 &\sigma_x^2 = v^2 \eta \sigma_o^2 \propto v^2 \eta =v \mathbb{E}d(v, \eta).
\end{flalign}
\end{subequations}

Here, $v\eta$ is the desired car-following distance. A new term $\sigma_o^2$ is introduced to represent the variance of stochastic car-following behavior propagated from the complex robotic uncertainties of AV operations, which is independent of variables related to vehicle control strategy such as speed $v$ and headway $\eta$.

\section{The Macroscopic Safety and Mobility}\label{sec:4}
We now shift our analysis from the microscopic scale of the fully autonomous traffic described by speed, headway, and inter-vehicle distance at an arbitrary fixed point to the macroscopic scale, represented by speed, density, and throughput or flow rate across roadway segment. As suggested by Eq~\eqref{eq:s_gen}, the maximum throughput becomes stochastic due to the possibility of collisions. This section then delineates the analytical derivation of the stochastic properties of maximum throughput in fully autonomous car-following scenarios.

\subsection{Measurement of traffic safety}
As the first step, we introduce a measurement of traffic safety, named \textit{collision rate}, using the stochastic car-following distances discussed above.

\subsubsection{Collision probability}
\begin{figure}[!ht]
 \centering
 \includegraphics[width=\textwidth]{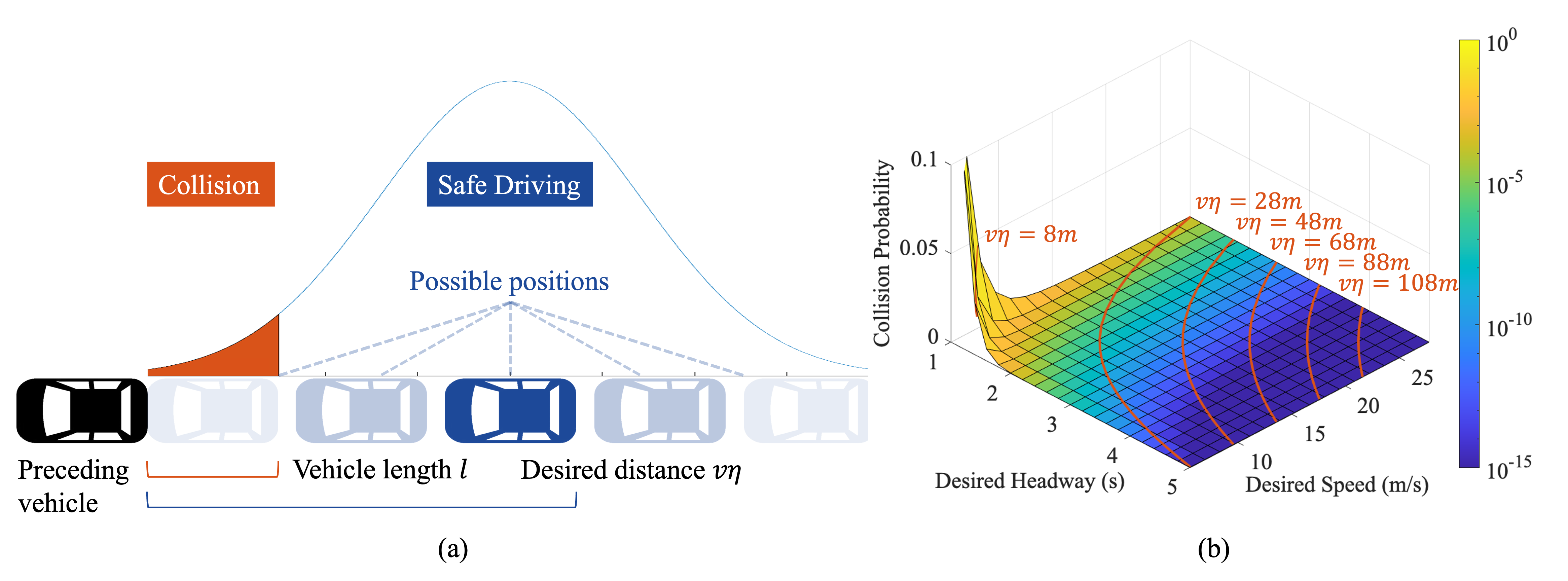}
 \caption{(a) Illustration of a rear-end collision and the corresponding collision probability under stochastic car-following distance; (b) Rear-end collision probability due to robotic uncertainty  with varying desired speed and headway.} 
 \label{fig:safety}
\end{figure}

 Although traffic safety is measured differently in the literature, the probability of collision is the most intuitive and widely-accepted indicator, especially in autonomous driving analysis \citep{de2021risk}. In homogeneous fully autonomous traffic, rear-end collisions occur when a bump-to-bump gap $d(v,\eta)$ becomes less than a vehicle's length $l$, as illustrated in Figure~\ref{fig:safety}(a). Since the stochastic car-following distance follows a Gaussian distribution in a statistical manner, the collision probability $p$ of a pair of AVs per time step could be expressed as a bi-variate function of speed $v$ and headway $\eta$:
\begin{flalign}
 p(v,\eta) = F_{d(v,\eta)}(l) = \int_{-\infty}^{l} \frac{1}{\sqrt{2\pi v^2 \eta \sigma_o^2}} exp\left(-\frac{(\omega -v\eta)^2}{2 v^2 \eta \sigma_o^2}\right) d \omega. \label{eq:pc}
\end{flalign}

Here, $F_{d(v,\eta)}(l)$ represents the cumulative probability of stochastic car-following distance being less than or equal to the vehicle length $l$. Its analytical form is obtained by integrating the Gaussian density function over $\omega$, which spans from $-\infty$ to $l$, the domain where a rear-end collision between the pair occurs. 

Figure~\ref{fig:safety}(b) demonstrates how collision probability varies with headway $\eta$ and speed $v$, assuming $\sigma_o=0.05 \text{ s}^{1/2}$ and $l=5$ m. The color shows the base-10 logarithm of the probability to emphasize its order of magnitude. Align with the analytical relationship demonstrated in Eq~\eqref{eq:pc}, the figure shows that the collision probability for a single pair of AVs per time step decreases with increasing speed for a fixed headway $\eta$. On the other hand, for a fixed speed, more conservative driving policies (i.e. larger headways) enhance traffic safety. In addition, the collision probability is more sensitive to changes in headway than to changes in speed. Especially at higher speeds, the collision probability shows only marginal differences. For example, the corresponding collision probabilities at speeds of $20$m/s ($72$ km/h) and $25$m/s ($90$km/h) are nearly identical. This asymmetry also implies the same desired car-following distance may not result in the same collision probability. As illustrated in the figure, five red lines are marked to present constant desired car-following distances, which are 8m, 28m, 48m, 68m, 88m, and 108m, respectively. The grids the lines across change drastically, indicating the collision probability increases. 

\subsubsection{Collision rate}
As outlined in the conceptual scenarios in Section~\ref{sec:2}, the macroscopic safety level can be evaluated by the collision rate of AVs per unit driving range and time in the transportation system. In car-following scenarios, the collision rate corresponds to the expected number of collisions occurring on an examined roadway segment per time step. Accordingly, it depends on the vehicle density $\rho$ on the roadway segment, the operational time step of AV operations $\tau$, and the collision probability of a single pair of AVs per time step $p(v,\eta)$, which is assumed to be independent across pairs. As a result, the collision rate $P$ for the longitudinal AV string along the roadway segment can be derived as a special case of Eq~\eqref{eq:P_gen}:
\begin{flalign}
 &P(v,\eta) = \frac{\tau L}{\tau \mathbb{E}d(v,\eta)} p(v,\eta) = \frac{L}{v \eta} \int_{-\infty}^{l} \frac{1}{\sqrt{2\pi v^2 \eta \sigma_o^2}} exp\left(-\frac{(\omega -v\eta)^2}{2 v^2 \eta \sigma_o^2}\right) d \omega. \label{eq:rc}
\end{flalign}

Here, $L$ refers to the length of the roadway segment, and $\mathbb{E}d(v,\eta)$ denotes the expectation of car-following distances with desired speed and headway. Accordingly, the vehicle density is expressed as $\frac{L}{\mathbb{E}d(v,\eta)}$.

\subsection{Traffic states and the associated capacities}
\begin{figure}[!ht]
 \centering
 \includegraphics[width=0.65\textwidth]{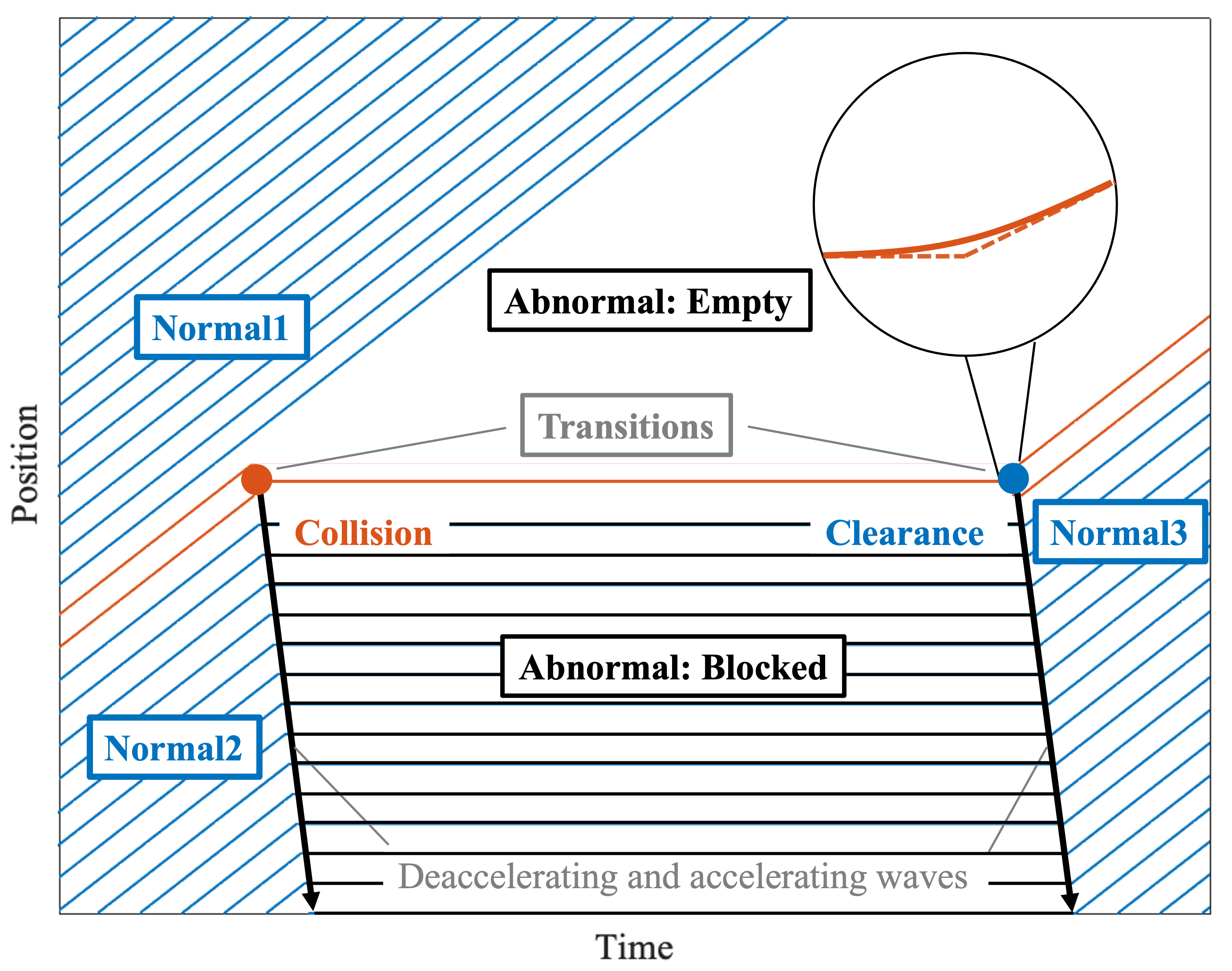}
 \caption{Traffic states consist of AVs' desired trajectories under normal driving and with collisions (traffic trajectories are oriented from the bottom-left to the top-right).}
 \label{fig:array}
\end{figure}

Traffic capacity usually refers to the maximum flow passing by a fixed location under stationary traffic conditions, excluding scenarios involving collisions. However, when stochastic car-following distance and resulting collisions are considered, a fixed location may witness additional traffic states. As shown in Figure~\ref{fig:array}, an occurred collision forces all following vehicles to stop completely, resulting in abnormal states both upstream (marked as blocked) and downstream (marked as empty). Furthermore, the transitions between normal and abnormal states also take some time due to vehicle acceleration and deceleration.

\subsubsection{Normal state}
In the normal state, the average lane capacity is derived as the reciprocal of the mean of stochastic headway $\eta$:
\begin{flalign}
 &s^+(v,\eta) = \frac{1}{\mathbb{E}\frac{d(v,\eta)}{v}} = \frac{1}{\eta}. \label{eq:s+}
\end{flalign}

The full capacity $s^+$ is the maximum attainable traffic flow rate under the safe condition that equals the reciprocal of the expectation of stochastic headway. It also serves as the upper bound of CIC, achieved when the collision rate approaches zero.

\subsubsection{Abnormal state (empty or blocked)} \label{sec:3_1_2}
Referring to Figure~\ref{fig:array}, once a rear-end collision occurs, its influence spreads both downstream and upstream. A downstream location of the collision becomes empty once the precedent vehicle in front of the collision drives past, ending the normal traffic state. The location will stay in the empty state until the complete clearance of the collision site, whose duration is usually called the total clearance time (TCT) \citep{smith2002forecasting} and is denoted as $T$ hereafter. Clearly, the maximum flow rate in the empty state is zero:
\begin{subequations} \label{eq:s-}
\begin{flalign}
 &s_e^-(v,\eta) = 0.
\end{flalign}

Then when the first vehicle after the clearance arrives, the empty state ends, and the normal state resumes. In the time-space diagram, all homogeneous AVs maintain the same desired speed, giving the empty state parallel wave boundaries. Thus, the time in the empty state for any locations downstream of the collision is equal to $T$.

Upstream locations, on the other hand, enter the blocked state as the vehicles passing by are forced to stop due to the collision ahead. The propagation from the normal state to the blocked state contributes to a shock-wave in the time-space diagram, squeezing the inter-vehicle distance of the car-following string, and the flow rate changes from $1/\eta$ to zero correspondingly:
\begin{flalign}
 &s_b^-(v,\eta) = 0.
\end{flalign}

After a period of collision clearance $T$, the road restores to normal gradually starting from the location of the collision. Essentially, the stopping and restoring shock-waves serve as the boundaries of the blocked state. The two shock-waves travel at the same speed, which can be expressed as: 
\begin{equation}
 c=\frac{q_n-q_b}{\rho_n-\rho_b}.
\end{equation}
Here, $q_n$ and $\rho_n$ denote the flow rate and density of the normal state, while $q_b$ and $\rho_b$ denote those of the abnormal (blocked) state. The two shock-waves are parallel and propagate from downstream to upstream. Consequently, irrespective of where it occurs, the abnormal state, including the empty and blocked states, persists for the same duration $T$ with zero throughput:
\begin{flalign}
 &s^-(v,\eta) = s_e^-(v,\eta) = s_b^-(v,\eta) = 0.
\end{flalign}
\end{subequations}
The zero throughput is the lower bound of CIC, achieved when the collision rate increases to one.

\subsubsection{Transitional State}
In reality, two additional transitional processes occur: the declaration of blocked AVs from desired speed $v$ to zero, and their acceleration from zero to $v$ after the clearance. However, these short-term transitions always accompany a collision and can therefore be considered as the margins of an abnormal (blocked) state. In addition, the duration of these two states is significant shorter than the total clearance time\footnote{Stopping and restoring usually happen within seconds, whereas the total clearance time could last more than half an hour, where evidence will be given in Section~\ref{sec:4_3_2}.}. In this context, the impact of these two transitions on calculating CIC can be ignored. Thus, we only consider normal and abnormal states in the subsequent analyses.

\subsection{Collision-inclusive capacity}
As introduced in Section~\ref{sec:2}, the expected lane capacity considering collisions caused by AVs' robotic uncertainties can be derived by a weighted average of traffic capacity in different states, i.e., $s = (1-\lambda) s^{+} + \lambda s^{-}$ (see Eq~\eqref{eq:s_gen}). We now introduce a semi-Markov process to describe the transitions and determine the corresponding weight $\lambda$. 

\subsubsection{The semi-Markov process}
\begin{figure}[!ht]
 \centering
 \includegraphics[width=0.85\textwidth]{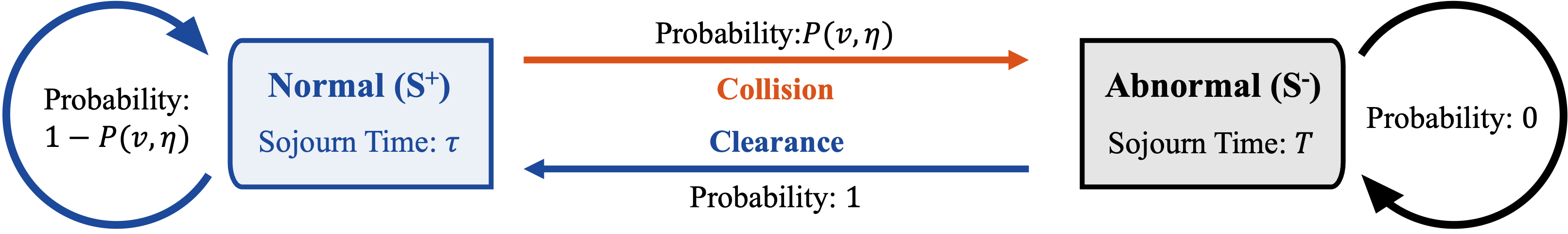}
 \caption{The semi-Markov process of traffic states for a single lane roadway segment.}
 \label{fig:stochastic_process}
\end{figure}

As illustrated in Figure~\ref{fig:array}, one collision can create three disjoint normal states on the two-dimensional space-time plan: the first represents the downstream traffic unaffected by the collision, the second depicts the upstream free flow before encountering the stopping shock-wave, and the third occurs to the right of the restoring shock-wave, indicating blocked traffic recovering to normal after the collision is cleared. Suppose the collision happens at time $0$ and position $x_c$ along the single-lane roadway segment of length $L$, Table~\ref{tab:indpdnt} summarizes the states each location will experience from time $0$, if no other collision happens. However, a second collision in the same lane may occur within one of the three normal traffic flows. Here, we focus on the case where a second collision happens only in the third normal state, making it independent of the first one. The analysis of second collisions in the other two normal states, which result into mutually is deferred to the extensions discussed in Section~\ref{sec:6}.
\begin{table}[htpb]
 \centering
 \caption{Time of three states for different locations on the roadway segment.}
 \label{tab:indpdnt}
 \begin{tabular*}{\tblwidth}{@{} L|LLL @{}}
 \toprule
 Location & Normal \textit{before collision} & Abnormal (Empty or Blocked) & Normal \textit{after clearance}\\
 \midrule
 $x\in [0, x_c)$ (downstream) & $\left[0, \frac{x_c-x}{v}\right)$ & $\left[\frac{x_c-x}{v}, \frac{x_c-x}{v}+T\right)$ & $\left[\frac{x_c-x}{v}+T, \sim \right)$ \\
 $x\in [x_c, L]$ (upstream) & $\left[0, \frac{x-x_c}{c}\right)$ & $\left[\frac{x-x_c}{c}, \frac{x-x_c}{c}+T\right)$ & $\left[\frac{x-x_c}{c}+T, \sim\right)$ \\
 \bottomrule
 \end{tabular*}
\end{table}

Since the stopping and restoring shock-waves are parallel, each point along the roadway segment will experience both the normal and abnormal states, with consistent durations for each state across all points, regardless of collision position. Furthermore, the dynamic changes in traffic states can be described by a semi-Markov process, as illustrated in Figure~\ref{fig:stochastic_process}. The normal state is associated with a sojourn time of $\tau$, while the abnormal state has a sojourn time of $T$. For each point, the transition probability from the normal state to the abnormal state equals the collision rate on the roadway segment, $P(v,\eta)$, while that of remaining in the normal state is $1-P(v,\eta)$. Meanwhile, the abnormal state will certainly transition back to the normal state after time $T$. In addition, the collision rate $P(v,\eta)$ is assumed to be extremely small, as collision among AVs in fully autonomous traffic are rare events given the maturity of AV technology\footnote{However, theoretically, this value could exceed one, as it represents the expected number of vehicles involved in a collision on the roadway segment at each time step.}. Therefore, the semi-Markov process governing traffic state changes remains mathematically well-defined.

Utilizing the semi-Markov process, we derive the weight $\lambda(v,\eta)$ from the following equations:
\begin{subequations}
\begin{flalign}
 &\lambda(v,\eta) = \frac{s(v,\eta)}{s^+(v,\eta)} \frac{T}{\tau} P(v,\eta), \label{eq:s_e_a}\\
 &s(v,\eta) = \left(1-\lambda(v,\eta)\right)s^+(v,\eta) + \lambda(v,\eta) s^-(v,\eta). \label{eq:s_e}
\end{flalign}

Solving the above equations leads to:
\begin{flalign}
 &\lambda(v,\eta) = \frac{1}{1+\frac{\tau}{T P(v,\eta)}}. \label{eq:lambda}
\end{flalign}
\end{subequations}

Together with Eqs~\eqref{eq:rc}, \eqref{eq:s+}, and \eqref{eq:s-}, the analytical form of CIC can then be derived as follows, which serves as a mobility measurement for the macroscopic fully autonomous traffic considering robotic uncertainties:
\begin{flalign}
 s(v,\eta) = \frac{s^+(v,\eta)}{1+\frac{T}{\tau}P(v,\eta)} = \frac{1}{\eta + \frac{T L}{\tau v} \int_{-\infty}^{l} \frac{1}{\sqrt{2\pi v^2 \eta \sigma_o^2}} exp\left(-\frac{(\omega -v\eta)^2}{2 v^2 \eta \sigma_o^2}\right) d \omega}. \label{eq:s_e_fin}
\end{flalign}

\subsubsection{Discussions} \label{sec:4_3_2}
\begin{figure}[htpb]
 \centering
 \includegraphics[width=\linewidth]{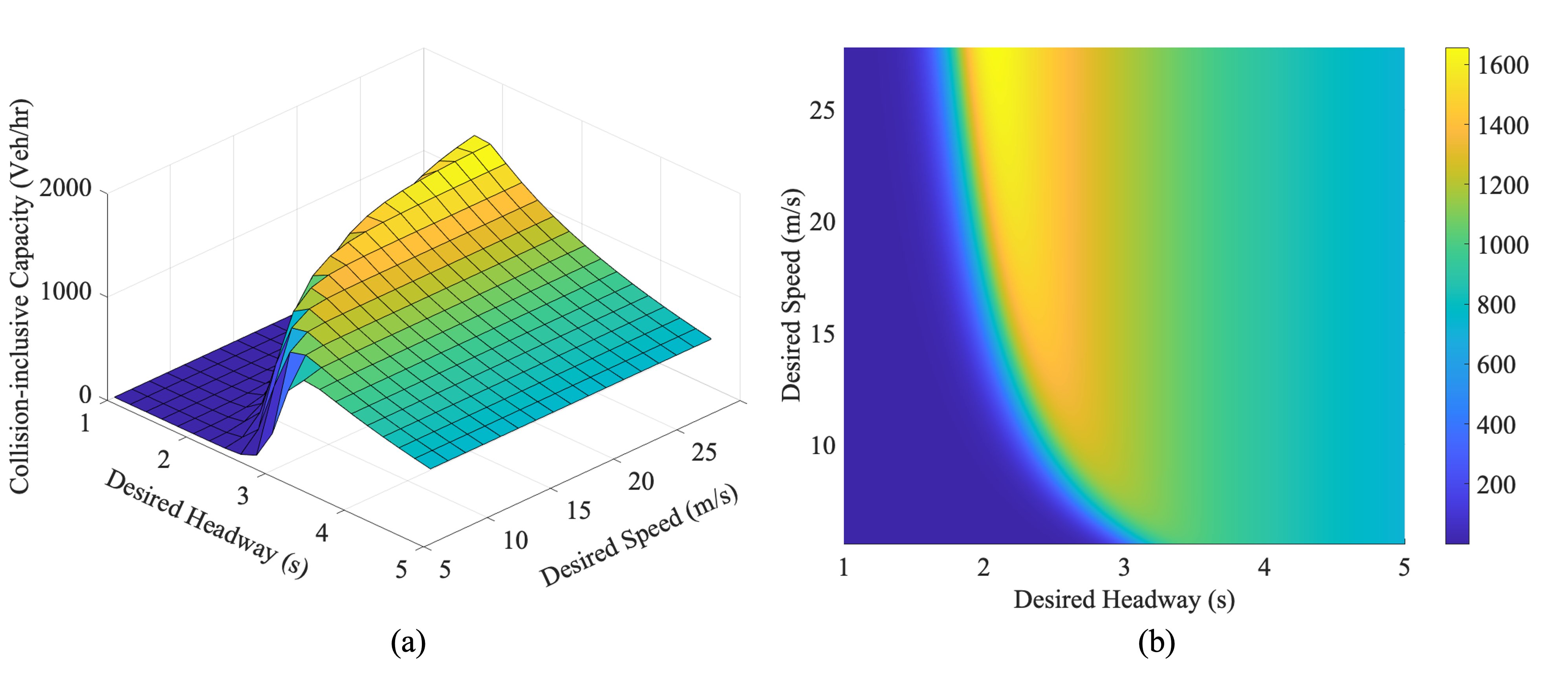}
 \caption{Relationship between macroscopic fully autonomous traffic mobility with desired speed and headway. (a) Three-dimensional representation of CIC as a function of speed and headway; (b) Two-dimensional projection of CIC onto the speed-headway plane.}
 \label{fig:capacity}
\end{figure}

Using the closed-form equation, we now analyze how CIC varies with speed and headway. The results are illustrated in Figure~\ref{fig:capacity}. Consistent with previous settings, we set $\sigma_o=0.05$, and $l=5$. Additionally, the roadway segment $L$ is set to be $5000$ meters and the time step $\tau$ is $0.1$s. The total clearance time (TCT) is assumed to be a linear function of speed, ranging from a minimum of $30$ minutes at $0$ km/h to a maximum of $60$ minutes at $120$ km/h. This assumption is based on empirical evidence that higher speeds can cause more severe collisions at high traffic flow, resulting in a longer clearance time \citep{christoforou2010vehicle}. Meanwhile, it is observed that traffic flow information does not contribute much to the accuracy of clearance duration predictions \citep{mihaita2019arterial}. Furthermore, HDV collision duration data in North Virginia \citep{dougald2016traffic} and San Francisco, USA, and Sydney, Australia \citep{grigorev2022incident} indicate that that TCT usually varies from half to an hour.

When the desired speed is fixed and headway increases, CIC initially rises sharply before gradually declining. Conversely, the capacity is only sensitive to specific speed ranges when the desired headway is roughly between $2$ s and $3$ s. Outside of this range, capacity remains largely unaffected by speed changes. Overall, CIC is more sensitive to variations at high speeds and small headways, where collisions are more likely. Nonetheless, the global maximum capacity also appears within this range.

\begin{figure}[!ht]
 \centering
 \includegraphics[width=0.65\linewidth]{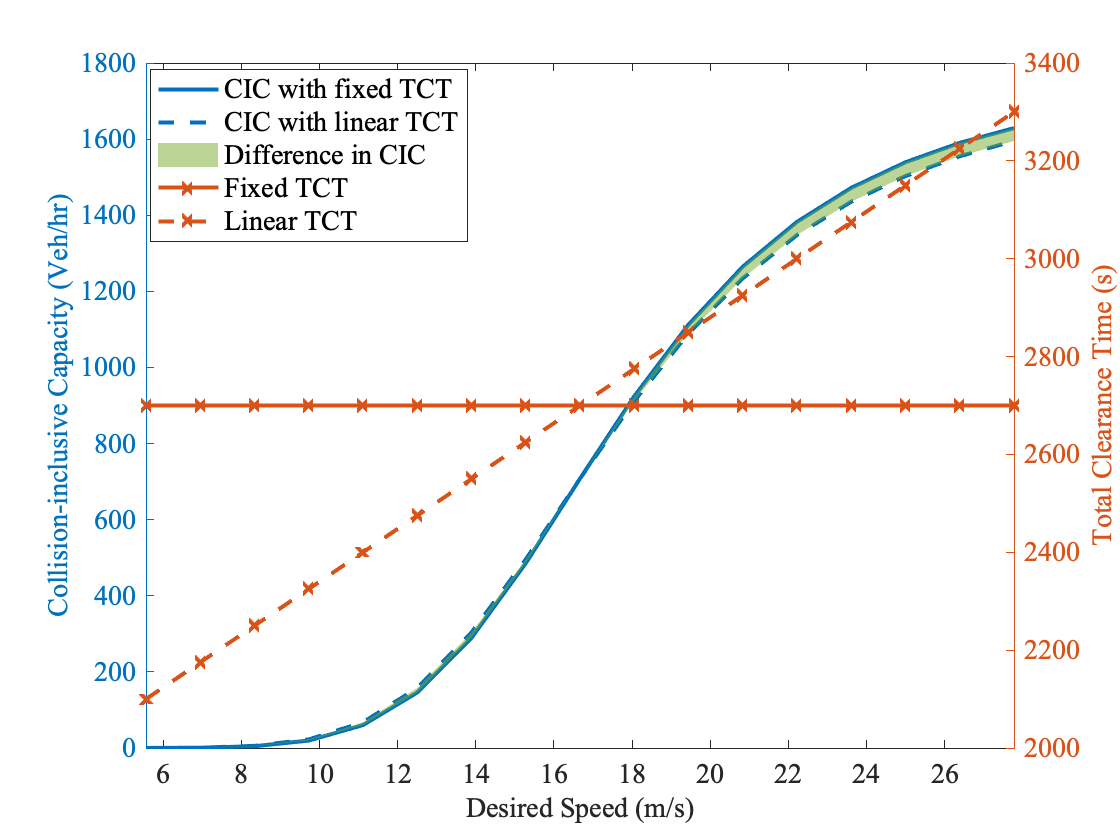}
 \caption{Relationship between collision-inclusive capacity and speed with a fixed headway under different types of TCTs.}
 \label{fig:TCT}
\end{figure}

In addition, as AV collision clearance has not been well explored in the literature, we also compare the assumed linear TCT with a fixed TCT of $45$ minutes at a typical headway of $\eta=2.0$ s, as shown in Figure~\ref{fig:TCT}. Generally, no significant differences between the two capacities are observed, suggesting that the independence of TCT from speed does not significantly affect CIC's properties. Upon closer inspection, only the capacity with the fixed TCT appears slightly larger at high speeds, as illustrated by the green shaded area in the figure.

\subsubsection{Comparison with collision-\textit{e}xclusive capacity}
\begin{figure}[t]
 \centering
 \includegraphics[width=0.65\textwidth]{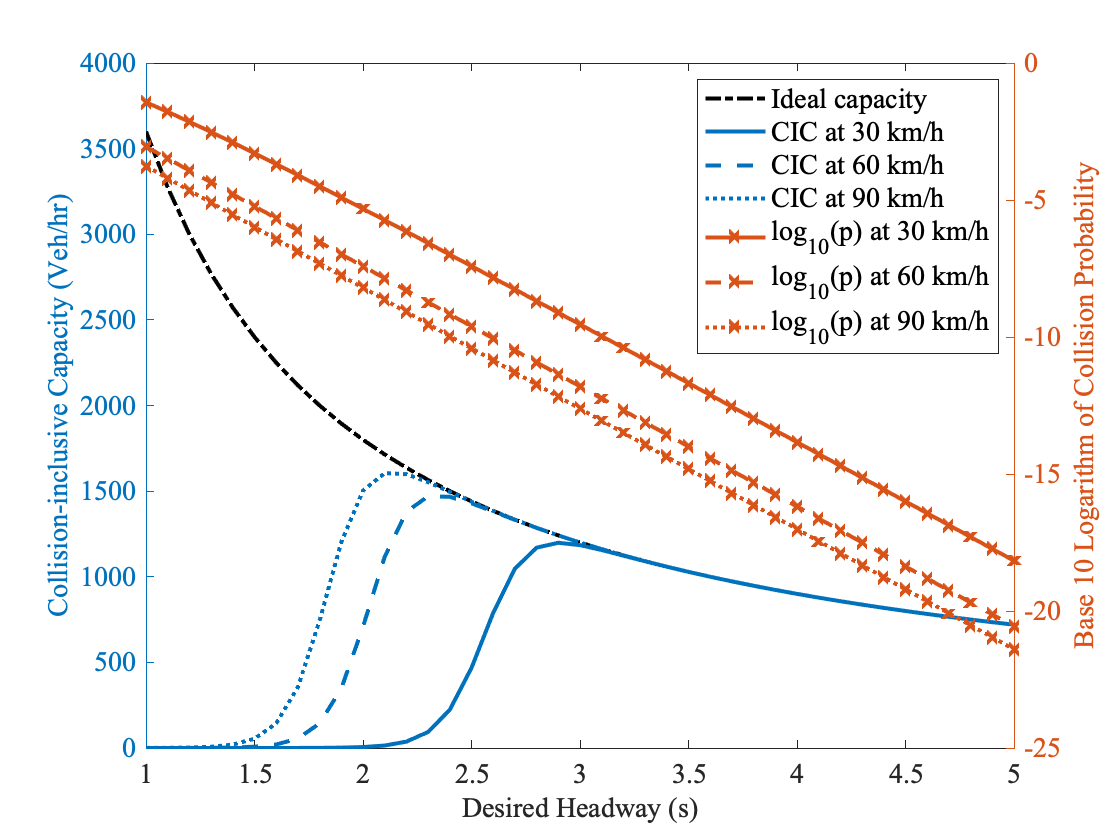}
 \caption{Comparison of our model results (blue curves) with perfect operation assumption (black curve) under different speeds and headway.}
 \label{fig:reference}
\end{figure}

Figure~\ref{fig:reference} compares CIC and the idealized capacity under perfect operation, demonstrating the macroscopic impact of AV robotic uncertainties on mobility. The AV capacity under perfect operation is given by $1/\eta$, indicating that it is only controlled by the headway. 

The comparison shows that when the desired headway is relatively large, both capacities are nearly identical. In this case, the capacity loss is attributed to the increasing spacing between vehicles, where robotic uncertainty does not significantly impact safety. However, as the headway decreases, the gap between the idealized capacity and the collision-inclusive one increases tremendously, until the latter drops to zero and remains there. This suggests that when the headway is small, the capacity loss from the idealized one is primarily due to the sharply increased collision rate under robotic uncertainty, which results in a longer average duration in the abnormal state. Furthermore, the two effects are asymmetric, where headway is a more sensitive variable at smaller values.

Unlike the ideal capacity, the optimal mobility indicated by CIC also depends on the speed. As shown by the orange lines, for a given headway, a smaller speed leads to a higher collision probability, resulting in a greater throughput loss. In the next section, we further investigate the optimal combinations of speed and headway to meet different traffic management requirements that prioritize mobility and safety, respectively.

\section{Suggestions on Design and Management}\label{sec:5}
In this section, we first conduct a series of sensitivity analyses to discuss the impact of $L$, $l$, and $\sigma_o$ on the safety and mobility performance of fully autonomous traffic, in order to give suggestions on the AV designs and adaptations on roads. While $\tau$ is suggested to be as short as possible, in line with \citet{li2022trade}'s suggestion that sensitivity should be as large as possible in the trade-off between safety, mobility, and stability. In addition, we further provide optimization over the two controllable variables $v$ and $\eta$ under different constraints of fully autonomous traffic to theoretically support traffic management and control.

\subsection{Suggestions on AV and road designs}
Among the parameters contributing to collision probability and CIC shown in Eqs~\eqref{eq:pc} and \eqref{eq:s_e_fin}, road length $L$ is an inherent property of road design. While vehicle length $l$ and precision $\sigma_o$ represent the performance of the AV design from both hardware and operating intelligence part.

\begin{figure}[!ht]
 \centering
 \includegraphics[width=\linewidth]{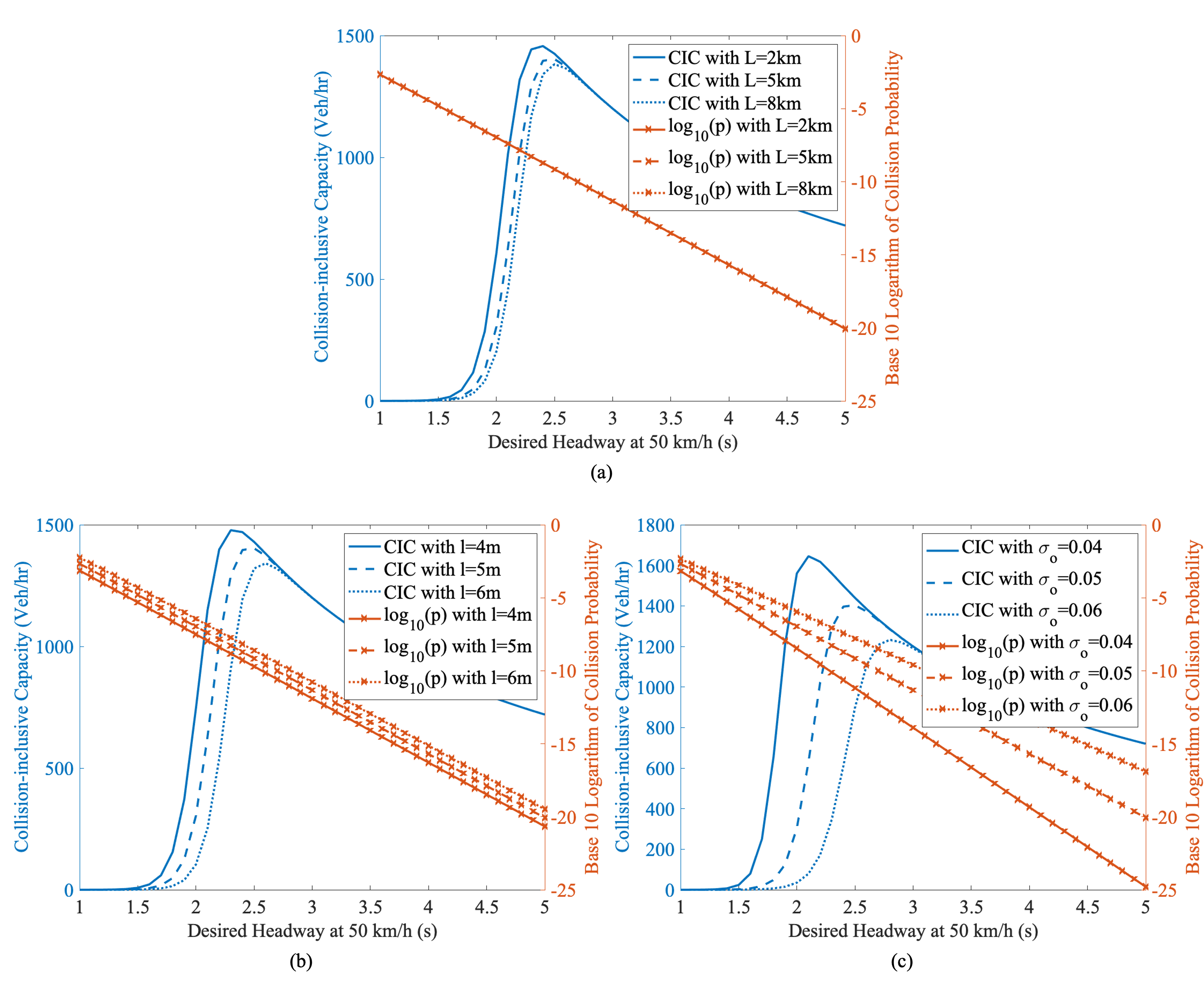}
 \caption{Relationship between collision-inclusive capacity and collision probability with desired headway at a fixed speed under different (a) road lengths, (b) vehicle lengths, and (c) independent robotic uncertainties.}
 \label{fig:sensitivity}
\end{figure}

\subsubsection*{Road length ($L$)}
The length of a roadway segment $L$ does not affect collision probability $p$ for each pair of AVs or the full capacity $s^+$, but it influences the collision rate $P$ and the associated CIC $s$. Specifically, a longer road segment carries more vehicles with the same density, resulting in a higher collision rate $P$ with the same collision probability $p$. Consequently, the transition probability from the normal state to the abnormal state becomes larger, resulting in a decline in the overall CIC.

Moreover, as shown in Figure~\ref{fig:sensitivity}(a), the decline is more obvious when the collision probability becomes large due to a more aggressive driving policy, i.e., a smaller $\eta$. Therefore, it is suggested that AV operations can be adaptive given different driving environments: When the roadway segment is longer, a more conservative driving strategy can be adopted using a relatively larger headway; While when the roadway segment is shorter, a more aggressive value of headway can be chosen.

\subsubsection*{Vehicle length ($l$)}
The length of a vehicle $l$ directly impacts collision probability. Under the same car-following distance (i.e. bump-to-bump gap), longer vehicles result in a shorter head-to-tail distance, increasing the collision probability. Meanwhile, the vehicle length $l$ has no influence on the full capacity $s^+$, but a negative impact on CIC, as shown in Figure~\ref{fig:sensitivity}(b).

Under the same headway, longer vehicles decrease CIC by increasing the collision rate. Alternatively, to maintain the same safety performance, longer vehicles require larger headway, so that to decrease the capacity. Therefore, compact vehicle design will become more favorable for fully autonomous traffic in the future.

\subsubsection*{Independent robotic uncertainty ($\sigma_o$)}
As the most important parameter to measure the independent robotic uncertainty of autonomous vehicles, $\sigma_o$ is essential and significant to both safety and mobility. As shown in Figure~\ref{fig:sensitivity}(c), with other parameters unchanged, a lower $\sigma_o$ would always infer vehicle movement with less uncertainty, so as to achieve fewer collisions and higher capacity in fully autonomous traffic. This property remains true regardless of the speed and the driving strategies being employed.

Similar to vehicle length, independent robotic uncertainty influences CIC by affecting collision probability (see Eq~\eqref{eq:pc}). More precise sensors and control with smaller $\sigma_o$ lead to safer traffic conditions under the same driving strategy. And with the same safety performance, smaller $\sigma_o$ allows smaller headway $\eta$ that contributes to higher traffic capacity and greater overall benefit. Therefore, it is always beneficial for AVs to achieve higher precision in their operation modules as it simultaneously improves traffic safety and mobility.

\subsection{Traffic management for fully autonomous traffic}
With the relationship between controllable variables and traffic safety and capacity, we can evaluate the benefits of AVs to the present transportation system and conduct optimization to further manage the fully autonomous traffic. 

While of paramount concern to AV manufacturers, safety does not exist in isolation as an optimization goal within the fully transportation systems. Stringently cautious strategies can indefinitely enhance security, albeit at the expense of a notable reduction in traffic capacity (see Figure~\ref{fig:reference}). Therefore, we consider two scenarios, one is trying to achieve the optimal system performance given the maximum allowable collision probability, and another is to find the optimal strategy with the minimum collision probability given that capacity satisfies traffic demand. Two stakeholders are involved in the management. The crucial parameters on the maximum allowable collision probability and the traffic demand are provided by transportation system, the society, and the government agency, while the speed of the AV string $v$ and headway $\eta$ are controlled by the AV manufacturers.

\subsubsection{Capacity improvement}
As there always exists a probability for collision, our first analysis is to identify the driving strategy that can maximize the system mobility under a maximum allowable collision probability $\hat{p}$. It leads to solving a constrained optimization problem:
\begin{flalign}
 \max_{v,\eta} \quad &s(v,\eta), \label{eq:opti_cap_vh} \\
 s.t. \quad &p(v,\eta) \le \hat{p}. \notag
\end{flalign}

We then adopt a two-stage approach to tackle this problem. In the first stage, we derive the optimal headway as a function of speed. In the second stage, we optimize the speed taking the analytical form of optimal headway into consideration. Notice that in the widely-circulated control and management methods for AV car-following strings \citep{lee2024traffic}, the headway is usually controlled for each AV and the speed is controlled for the string as a whole.\\ \hspace*{\fill} \\

\noindent\textbf{Stage One: Optimal headway with respect to a given speed}\\ 
In the first stage, we propose two lemmas to demonstrate the analytical form of optimal headway $\eta^*$.
\begin{lm}\label{lm:1}
  For any given maximum allowable collision probability $\hat{p}$ and fixed speed $v$, there exists a unique $\hat{\eta}_v$ such that $p_v(\eta) \le \hat{p} \iff \eta \ge \hat{\eta}_v$. Furthermore, $\hat{\eta}_v$ satisfies $p_v(\hat{\eta}_v)=\hat{p}$.
\end{lm}
\begin{proof}
Notice that $F_{d(v,\eta)}(l)=\textit{Prob} (v\eta +\epsilon_{x} \leq l)$. This allows us to define $H_v(\eta;l)$ as a function of $\eta$, given by:
\begin{flalign}
 H_v(\eta;l)=\textit{Prob} (\epsilon_x\leq l-v\eta ).
\end{flalign}

Clearly, $H_v(\eta;l)$ is a monotonically decreasing function of $\eta$. Therefore, $\eta$ and $p_v(\eta)$ a bijective mapping where $p_v(\eta)$ decreases with the increase of $\eta$.
\end{proof}

\begin{lm}\label{lm:2}
 Given speed $v$, the optimal headway $\eta^{\dag}$ of the optimization problem~\eqref{eq:opti_cap_vh} is always achievable and is a function of $v$.
\end{lm}

\begin{proof}
When AVs' operational speed is fixed at $v$, the original problem~\eqref{eq:opti_cap_vh} can be formulated as:
\begin{subequations}
\begin{flalign}
  \min_{\eta} \quad &\frac{1}{s_v(\eta)}=\eta + \frac{TL}{v\tau}p_v(\eta), \label{eq:opti_cap_h} \\
  s.t. \quad &\eta \ge \hat{\eta}_v, \notag
\end{flalign}
where $s_v(\eta)$ denotes the function of CIC under fixed speed $v$, and $\hat{\eta}_v$ is determined according to Lemma~\ref{lm:1}.

The first order condition of the reformulated problem suggests that:
\begin{flalign}
  1+\frac{TL}{v\tau}p'_v(\eta^*)=0, \label{eq: glopt_con}
\end{flalign}
where $\eta^*_v$ is the optimal solution of the underlying unconstrained program, and $p'_v(\eta^*_v)$ denotes the first order derivative, that is, $\frac{d p_v(\eta)}{d \eta}|_{\eta=\eta^*_v}$. As $p_v(\eta)$ is a monotonically decreasing function of $\eta$, it is theoretically possible that a $\eta^*_v\geq 0$ exists such that $p'_v(\eta^*_v)=-\frac{v\tau}{TL}$. Then, if $\forall \eta \geq \hat{\eta}_v$, $-\frac{v\tau}{TL}<p'_v(\eta)<0$, the objective is monotonically increasing with $\eta$, making the optimal headway being $\hat{\eta}_v$. If $\eta^*_v\geq \hat{\eta}_v$ and $p'_v(\eta^*_v)=-\frac{v\tau}{TL}$, $\eta^*_v$ is the optimal headway. In conclusion, the optimal headway of problem~\eqref{eq:opti_cap_h} can be denoted as:
  \begin{flalign}
   &\eta^{\dag}(v) = \max\{\hat{\eta}_v, \eta^*_v\}. \label{eq:optimal_hh}
  \end{flalign} 
\end{subequations}
\end{proof}

\begin{figure}[!ht]
 \centering
 \includegraphics[width=0.65\textwidth]{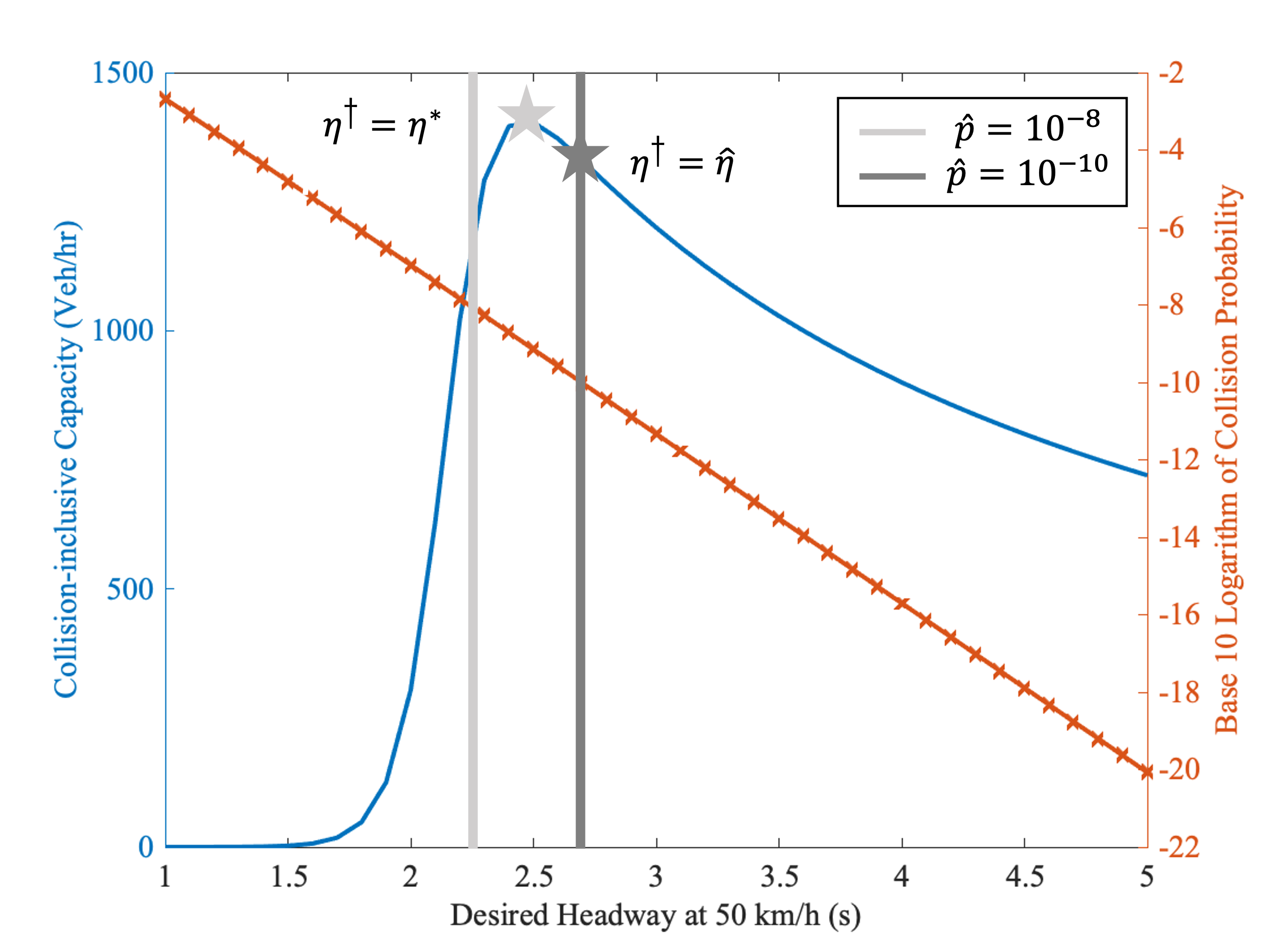}
 \caption{The change of locally optimal headway according to different safety constraints The blue curve represents the CIC. The red line depicts the collision probability on the logarithmic scale. The light gray star indicates the optimal headway for $\hat{p}=10^{-8}$, and the dark gray star marks the optimal headway for $\hat{p}=10^{-10}$.} 
 \label{fig:lcoal_optimal}
\end{figure}

Figure~\ref{fig:lcoal_optimal} illustrates the two possible optimal headway under Gaussian car-following distance assumption. When collision probability is confined to be under $10^{-10}$, the safety constraint is binding so that the optimal headway is given by $\hat{\eta}$. When the maximum allowable collision probability is relaxed to $10^{-8}$, the global optimal headway $\eta^*$ is achieved.

\noindent\textbf{Stage Two: Optimal speed for traffic mobility}\\
To evaluate the optimal speed that maximizes CIC, we first examine the relationship between speed and headway under the same level of collision probability. 
\begin{lm}\label{lm:3}
  When the collision probability remains constant, an increase in $v$ results in a decrease in $\eta$, and vice versa. 
\end{lm}
\begin{proof}
  From Lemma~\ref{lm:1}, we know that $p(v,\eta)$ is monotonically decreasing with $\eta$. Using the same logic, we denote:
  \begin{subequations}
  \begin{flalign}
    G_{\eta}(v;l)=\textit{Prob}(\epsilon_x\leq l-v\eta)=H_v(\eta;l),
  \end{flalign}
  which is a monotonically decreasing function of $v$, that is, $\frac{\partial p}{\partial v}<0$. Notice that:
  \begin{flalign}
    d p(v,\eta)=\frac{\partial p}{\partial v} dv +\frac{\partial p}{\partial \eta} d\eta =0 \Rightarrow
    \frac{d\eta}{dv}=-\frac{\partial p}{\partial v}/\frac{\partial p}{\partial\eta}.
  \end{flalign}
  \end{subequations}
  
  Since $\frac{\partial p}{\partial v}<0$, $\frac{\partial p}{\partial \eta}<0$, then $\frac{d\eta}{dv}$<0, suggesting the headway decreases with the increase of speed to maintain the same probability of collision. The proof is concluded.
\end{proof}

\begin{prop}\label{prop:1}
 The CIC under optimal headway $\eta^{\dag}$ is monotonically increasing with speed $v$. 
\end{prop}

\begin{proof}
For any pair of $v_1 < v_2$, the optimal headways are $\eta^{\dag}(v_1)$ and $\eta^{\dag}(v_2)$. We now introduce an augmented variable $\eta_r$, which is given as follows:
\begin{subequations}
\begin{flalign}
 &p(v_2, \eta_r) = p(v_1,\eta^{\dag}(v_1)). \label{eq:ref}
\end{flalign}

Given that $v_1 < v_2$, Lemma~\ref{lm:3} indicates that $\eta^{\dag}(v_1)>\eta_r$. Therefore,
\begin{flalign}
 s_{\eta^\dag}(v_2) &\ge s(v_2,\eta_r) \notag\\
 &= \frac{1}{\eta_r + \frac{T L}{\tau v_2} p(v_2,\eta_r)} = \frac{1}{\eta_r + \frac{T L}{\tau v_2} p(v_1,\eta^{\dag}(v_1))} \notag\\
 &> \frac{1}{\eta^{\dag}(v_1) + \frac{T L}{\tau v_1} p(v_1,\eta^{\dag}(v_1))} \notag\\
 &= s^*(v_1).
\end{flalign}
\end{subequations}
Here, the first inequality holds due to the optimality of $ s_{\eta^\dag}(v_2)$. The following equality is derived by definition, while the second equality is based on the assumption in Eq~\eqref{eq:ref}. Then the last inequality stands since $\eta^{\dag}(v_1)>\eta_r$. The inequality implies that CIC under speed $v_2$ with optimal headway $\eta^{\dag}(v_2)$ is greater than that under speed $v_1$ with optimal headway $\eta^{\dag}(v_1)$. Since $v_2$ and $v_1$ are an $v_2>v_1$, the proof is concluded.
\end{proof}
\noindent \textbf{Remark:} The lemmas and propositions hold so long as $F_{d(v,\eta)}(l)$ is a well-defined cumulative distribution function. Consequently, the stochastic car-following distance can follow any distribution, not limited to the Gaussian distribution. However, if it follows, $\hat{\eta}$ can be obtained from $\hat{p}$ by Gaussian distribution lookup table. The corresponding formulation of CIC can be explicitly stated:
\begin{flalign}
 & s_{\eta^{\dag}}(v) = \left({\eta^{\dag}(v) + \frac{T L}{\tau v} \int_{-\infty}^{l} \frac{1}{\sqrt{2\pi v^2 \eta^{\dag}(v) \sigma_o^2}} exp\left(-\frac{(\omega -v\eta^{\dag}(v))^2}{2 v^2 \eta^{\dag}(v) \sigma_o^2}\right) d \omega}\right)^{-1}. \label{eq:s_e_v}
\end{flalign}

\begin{figure}[htpb]
 \centering
 \includegraphics[width=0.7\textwidth]{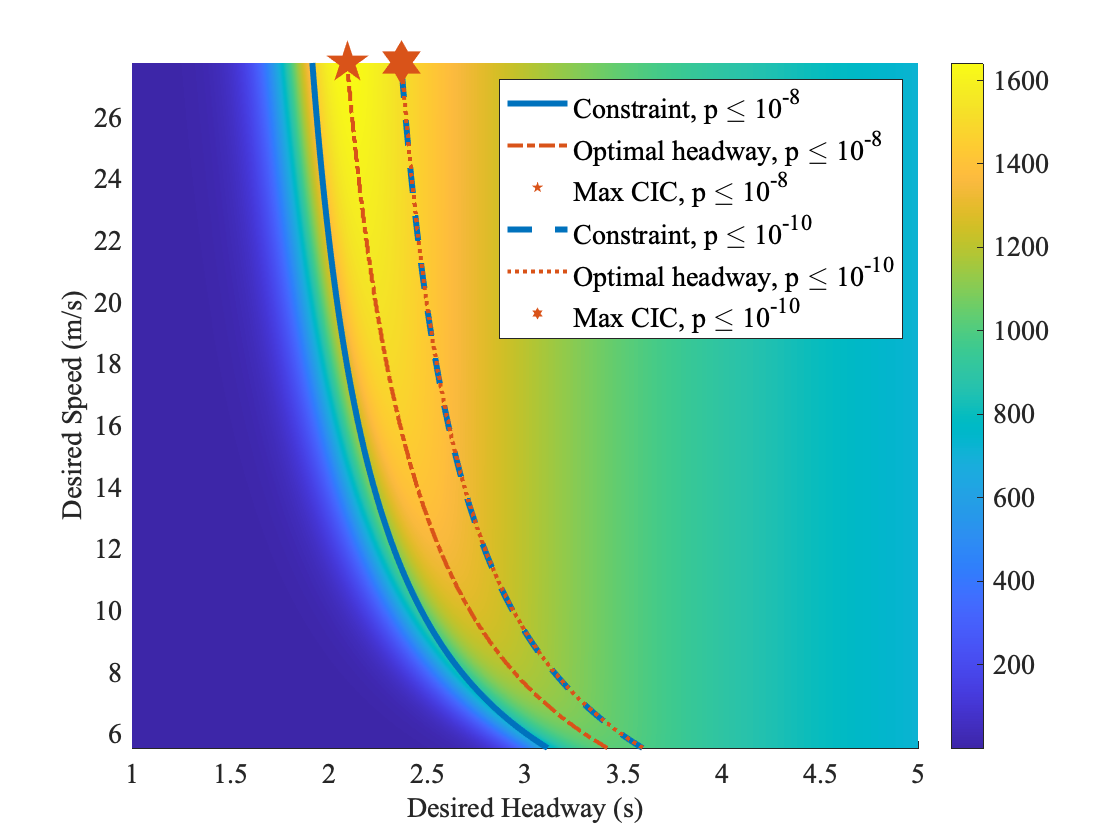}
 \caption{Impact of safety constraints on the optimal CIC under varying speed. Contours show the CIC across speed–headway combinations. Blue curves Lines represent safety constraints and corresponding optimal headway for $\hat{p} \leq 10^{-8}$ and $\hat{p} \leq 10^{-10}$ with varying speed. Stars mark the maximum CIC under each constraint.} \label{fig:global_optimal}
\end{figure}

Figure~\ref{fig:global_optimal} presents the variation of CIC with respect to the desired speed and headway, presented as a two-dimensional diagram where the color gradient represents the value of CIC. The blue solid line represents a milder safety constraint with $p\leq 10^{-8}$. Under this condition, the optimal headway for each speed always achieves the global optimal headway ($\eta^*_v$),indicated by the dashed red line. In contrast, the blue dashed line corresponds to a stricter safety constraint, where $p\leq 10^{-10}$. For this scenario, the optimal headway per each speed coincides with the value that makes the safety constraint binding. 

In accordance with Proposition~\ref{prop:1}, the optimal speeds for both scenarios are achieved at the highest speeds. While a larger speed always benefits CIC theoretically, practical speed limits are often imposed by road conditions such as curvature, slope, unevenness, etc. In addition, a higher speed with the corresponding optimal headway may not always result in a larger desired car-following distance. As can be observed in Figure~\ref{fig:global_optimal}, under the Gaussian distribution assumption, the optimal headway decreases as speed increases, leaving the product of speed and headway, i.e., the desired car-following distance, indeterminate. Therefore, it is recommended the AV string to drive as fast as possible only within reasonable limits, aligning with the intuitive goal of improving traffic capacity.

\subsubsection{Safety enhancement}
Nonetheless, when traffic demand is small, aggressive driving policies for high capacity are unnecessary. Instead, as long as the capacity can meet traffic needs, the driving strategy should be adjusted to reduce the collision probability to the greatest extent possible. Again, this management strategy can be represented by a constrained optimization problem:
\begin{flalign}
 \min_{v,\eta} \quad &p(v,\eta), \label{eq:opti_safe_vh} \\
 s.t. \quad &s(v,\eta) \ge \hat{s}.\notag
\end{flalign}

To solve this problem analytically, we again adopt the previous two-stage approach.\\

\noindent\textbf{Stage One: Optimal headway with respect to given speed}

\begin{lm}\label{lm:4}
If the optimization problem~\eqref{eq:opti_safe_vh} is feasible for a specified speed $v$, then there exists a critical headway $\eta^r_v$, defined as the maximum headway $\eta$ for which the capacity requirement $s(v,\eta) \ge \hat{s}$ holds. Furthermore, $\eta^r_v$ minimizes the collision probability $p(v,\eta^r_v)$ for the specified speed $v$. Mathematically, we define $\eta^r_v =\max \{\eta|s(v,\eta) \ge \hat{s}\}$, and $p(v,\eta^r_v)=\min_{\eta}p(v,\eta^r).$
\end{lm}

\begin{proof}
 According to the assumption, we know that for the given speed $v$, the set $\{\eta|s(v,\eta)\geq \hat{s}\}$ non-empty. 
 
 We now claim that the CIC cannot increase infinitely with respect to $\eta$, which can be easily shown by contradiction. If not, as $\eta$ approaches infinity, we have: 
 \begin{subequations}
 \begin{flalign}
  \lim_{\eta \rightarrow +\infty} s=\frac{1}{\eta+\frac{TL}{\tau v}p(v,\eta)}=\frac{1}{+\infty+\frac{TL}{\tau v}p(v,\eta)}=+\infty,
\end{flalign}
that implies:
\begin{flalign}
  \lim_{\eta \rightarrow \infty}\frac{1}{s}=+\infty+\frac{TL}{\tau v}p(v,\eta)=0,
\end{flalign}
which further suggests:
\begin{flalign}
  \lim_{\eta\rightarrow +\infty}\frac{TL}{\tau v}p(v,\eta)=-\infty.
\end{flalign}

It then leads to the conclusion that:
\begin{flalign}
  \lim_{\eta\rightarrow +\infty}p(v,\eta)=-\infty \times \frac{\tau v}{TL}=-\infty.
\end{flalign}
\end{subequations}

However, since $p(v,\eta)$ defines the probability function, it must always be a positive value less or equal to one, which contradict to the assumption.

Since the CIC cannot increase infinitely with respect to $\eta$, there exits a critical headway $\eta_v^r=\max \{\eta|s(v,\eta)\geq \hat{s}\}$. As suggested by Lemma~\ref{lm:1}, the collision probability under any specified speed decreases with the increase of $\eta$, i.e., $p'_v(\eta)<0$, then $\eta_v^r$ minimizes $p(v,\eta)$ for the specific speed $v$. The proof is concluded.
\end{proof}
This lemma suggests that under a fixed speed $v$, the highest headway that satisfies the capacity requirement also leads to the lowest probability of collision.\\

\noindent\textbf{Stage Two: Optimal speed for safety enhancement}

\begin{prop}\label{prop:2}
 The collision probability under optimal headway $p(v,\eta^r_v)$ is a decreasing function of speed $v$.
\end{prop}

\begin{proof}
For any pair of $v_1 < v_2$, suppose the the corresponding optimal headway for problem~\eqref{eq:opti_safe_vh} are $\eta^r_{v_1}$ and $\eta^r_{v_2}$. Since $s(v_1, \eta^r_{v_1})=s(v_2, \eta^r_{v_2})=\hat{s}$, we have:
\begin{subequations}
\begin{flalign}
 \frac{1}{s(v_2, \eta^r_{v_1})}=\eta^r_{v_1} + \frac{T L}{\tau v_2} p(v_2,\eta^r_{v_1}) &< \eta^r_{v_1} + \frac{T L}{\tau v_1} p(v_1,\eta^r_{v_1}) = \frac{1}{\hat{s}}=\eta^r_{v_2} + \frac{T L}{\tau v_2} p(v_2,\eta^r_{v_2}).
\end{flalign}

The first inequality holds since the collision probability monotonically decreases with the increase of speed at the same headway. Given that Therefore, we derive that:
\begin{flalign}
  s(v_2,\eta^r_{v_1}) > s(v_2,\eta^r_{v_2}). \label{eq:prop2_com}
\end{flalign} 

According to Lemma~\ref{lm:4}, $\eta^r_{v_2}$ is the highest $\eta$ that makes $s(v_2,\eta^r_{v_2})\geq \hat{s}$. Therefore, $\forall \eta> \eta^r_{v_2}$, $ s(v_2,\eta)<\hat{s}$. In this context, we conclude from Eq~\eqref{eq:prop2_com} that:
\begin{flalign}
  \eta^r_{v_1}<\eta^r_{v_2}.
\end{flalign} 

Meanwhile, since $v_1<v_2$ by assumption, we further have:
\begin{flalign}
 p(v_2,\eta^r_{v_2})<p(v_2,\eta^r_{v_1})<p(v_1,\eta^r{v_1}).
\end{flalign}
\end{subequations}

The proof is concluded.
\end{proof}

When the objective is collision minimization, Proposition~\ref{prop:2} indicates that both the optimal speed and headway should be maximized so long as the capacity constraint is met. Accordingly, this maximization results in a larger desired car-following distance. This result contrasts with the case of capacity improvement, where the desired car-following distance under the optimality may not be the largest.

\begin{figure}[htpb]
 \centering
 \includegraphics[width=0.7\textwidth]{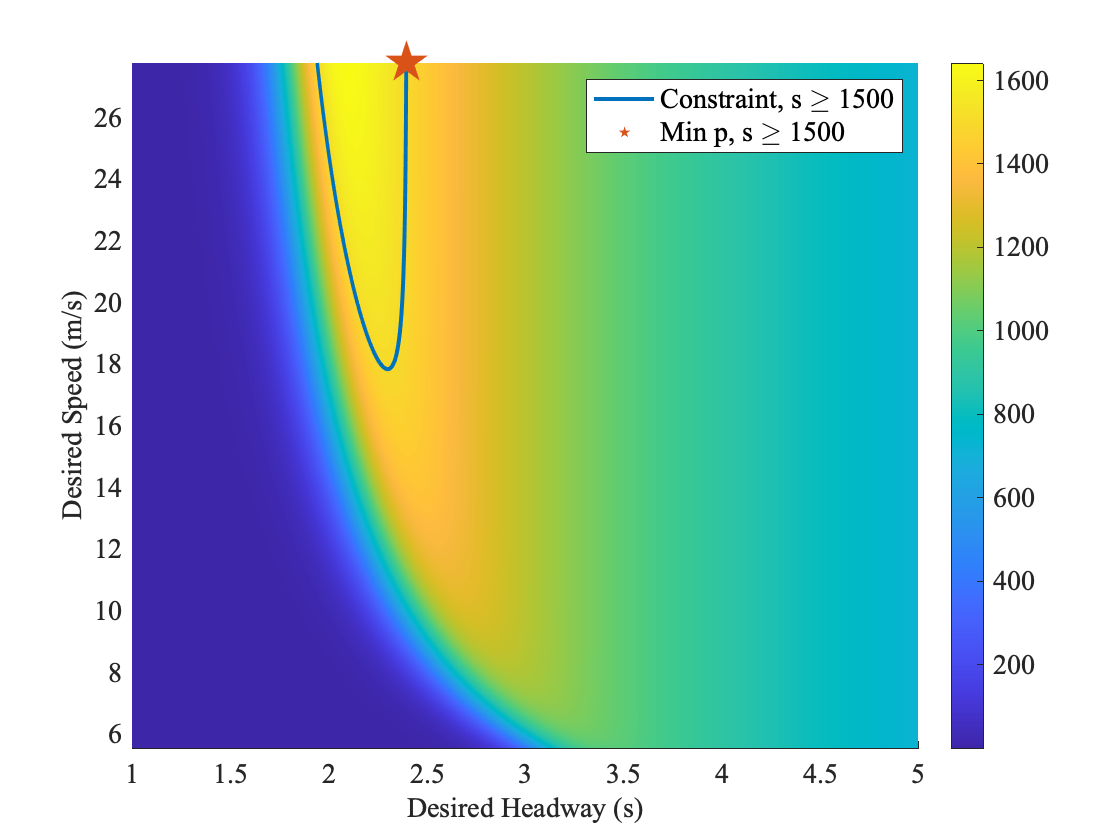}
 \caption{Safety enhancement with a demand constraint. The blue line confines the feasible region under demand constraint. The star symbol marks the corresponding CIC when collision probability is minimized.} \label{fig:global_optimal_P}
\end{figure}

Figure~\ref{fig:global_optimal_P} provides an example for safety enhancement with Gaussian distribution assumption. The feasible region is circled by the blue line. The optimal policy is at the upper right corner of the region, with the maximum feasible speed and maximum feasible headway. In summary, for the macroscopic traffic management for fully autonomous traffic, setting a high speed is in general advantageous, whether the objective is capacity improvement or safety enhancement. However, the optimal headway should be adjusted based on the prioritized objective. When safety is the primary concern, the corresponding headway is typically greater than that required for capacity improvement.

\section{Extensions} \label{sec:6}
Building on the previous analysis which primarily focused on single-lane scenarios and independent collisions, this section extends the model framework to account for non-independent collisions and multi-lane roads, enabling an evaluation of its generalization capability.

\subsection{Overlapping collisions}

\begin{figure}[!ht]
 \centering
 \includegraphics[width=0.85 \textwidth]{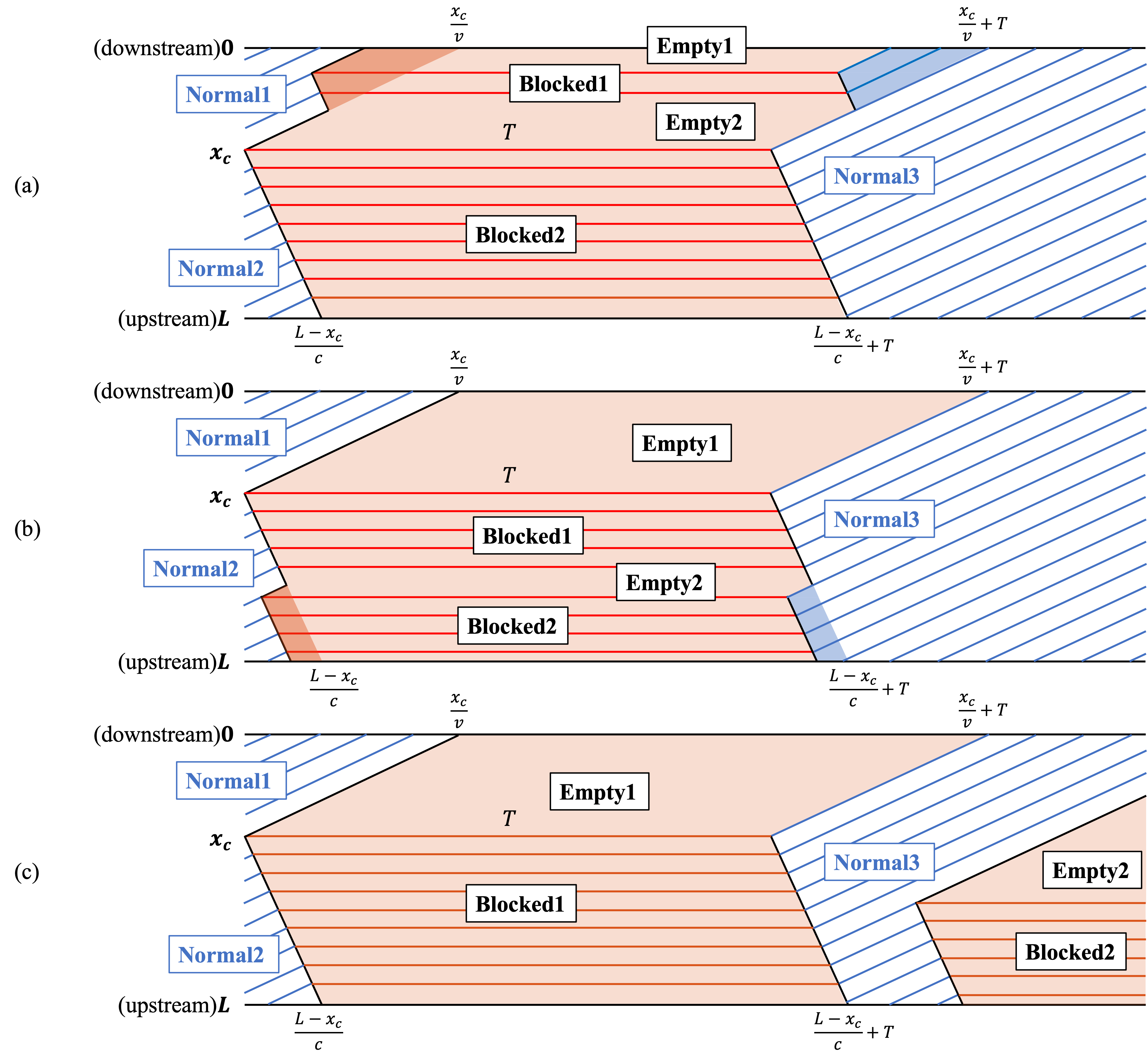}
 \caption{Desired traffic flows of the situation where a second independent collision happens in (a) "Normal1", (b) "Normal2", and (c) "Normal3".} 
 \label{fig:overlapping}
\end{figure}

Figure~\ref{fig:overlapping} overviews the locations of a second possible collision when the first collision happens. As can be seen, the previous analysis focuses on independent collisions is aligned with the demonstration in Figure~\ref{fig:overlapping}(c), where the second possible collision happens in the ``Normal 3'' regime. Next, we move to the scenarios when the second possible collision happens in ``Normal1'' or ``Normal2'', which corresponds to Figure~\ref{fig:overlapping}(a) and Figure~\ref{fig:overlapping}(b), respectively.

As shown in Figure~\ref{fig:overlapping}(a), ``Normal1'' is located downstream of the original collision, allowing its vehicles to move freely to the end of the road at location $0$, unaffected by the original collision. If a second collision occurs in ``Normal 1'', it halts the upstream vehicles within the same regime as illustrated by the shaded red area. The abnormal state will be recovered to the new normal state after a period of $T$, with the corresponding normal operations depicted by the shaded blue area. The two shaded areas are of equal size, as represent the movements of the same set of vehicles over the same amount of length. Therefore, compared to the case with only one collision, the second collision in ``Normal1'' merely shifts the flow that would have traversed ``Normal1'' to "Normal3". Although this creates a different time-space diagram, the shift does not alter the road capacity dynamics captured by the previous semi-Markov process.

We now turn to the case that a second collision occurs in ``Normal2''. A second collision in this regime only affects upstream vehicles alongside the road to its entrance at location $L$. Vehicles affected by the second collision will be blocked earlier than the original collision, as illustrated by the shaded red area in Figure~\ref{fig:overlapping}(b). However, they will also recover earlier, catching up with the previous cars after the collision is cleared, and continue to form a car-following string, as depicted by the shaded blue area in Figure~\ref{fig:overlapping}(b). Similar to the previous case, the road capacity dynamics can still be depicted by the previous semi-Markov process.

In conclusion, the location of the second collision before the first one is cleared does not affect the nature of the capacity dynamics at each single location along the roadway, whose stationary property is captured by CIC introduced in the last section. Nonetheless, when considering the traffic throughput within a limited amount of time, the location of the second collision matters. In other words, the impacts of overlapping collisions and independent collisions are different. When a limited study period is assumed, we examine the expected value of traffic throughput of the roadway segment as the corresponding measure. Here, we provide the general form of CIC for one lane under overlapping collisions, $s_c(v,\eta)$, as follows:
\begin{subequations}
\begin{flalign}
 s_c(v,\eta) &= \sum_{i=0}^{\infty} p_{c,i}(v,\eta) s_{c,i}(v,\eta).
\end{flalign}

Here, $p_{c,i}$ shows the probability that the $i$th collision occurs, while $s_{c,i}$ indicates the associated throughput. In this context, $p_{c,0}$ represents the probability that no collision happens in the study period, and $s_{c,0}=s^+$. When the study period is assumed to be one hour, i.e., $H=1$ hr $=3600$ s,  the specific form with at most two overlapping collisions is given by:
\begin{flalign}
 s_c(v,\eta) &\approx p_{c,0}(v,\eta) s_{c,0}(v,\eta) + p_{c,1}(v,\eta) s_{c,1}(v,\eta) + p_{c,2}(v,\eta) s_{c,2}(v,\eta) \label{eq:cap2},
\end{flalign}
in which the first two terms are given as follows: 
\begin{flalign}
 p_{c,0}(v,\eta) s_{c,0}(v,\eta) &= \left(1-p(v,\eta)\right)^{\frac{H}{\tau} \frac{L}{\eta v}} s^{+}(v, \eta) \notag \\
  &\approx \left( 1 - \frac{H}{\tau} \frac{L}{\eta v} p(v,\eta) + \frac{\frac{H}{\tau} \frac{L}{\eta v}\left(\frac{H}{\tau} \frac{L}{\eta v}-1\right)}{2} p(v,\eta)^2 \right) s^{+}(\eta); \label{eq:cap2collision_1}
\end{flalign}
\begin{flalign}
 p_{c,1}(v,\eta) s_{c,1}(v,\eta) &= \frac{H}{\tau} \frac{L}{\eta v} p(v,\eta) \left(1-p(v,\eta)\right)^{\frac{H}{\tau} \frac{L}{\eta v} -1} \left(1-\frac{T}{H}\right) s^{+}(v,\eta) \notag \\
 &\approx \frac{H}{\tau} \frac{L}{\eta v} p(v,\eta)\left( 1 - \frac{H}{\tau} \frac{L}{\eta v} p(v,\eta) + \frac{\frac{H}{\tau} \frac{L}{\eta v}\left(\frac{H}{\tau} \frac{L}{\eta v}-1\right)}{2} p(v,\eta)^2 \right)\left(1-\frac{T}{H}\right) s^{+}(v,\eta)\notag\\
 &\approx \left( 1 - \left(\frac{H}{\tau} \frac{L}{\eta v} -1\right) p(v,\eta) \right) \frac{H}{\tau} \frac{L}{\eta v} p(v,\eta) \left(1-\frac{T}{H}\right) s^{+}(v,\eta). \label{eq:cap2collision_2}
\end{flalign}

Here, the equalities in both equations holds due to the independence nature of the occurrence of collisions at different locations within time step. Specifically, the probability that no collisions occur during the study period, denoted by $p_{c,0}(v,\eta)$, is given by to $\left(1-p(v,\eta)\right)^{\frac{H}{\tau} \frac{L}{\eta v}}$, and the associated capacity under this no-collision scenario corresponds to the full capacity $s^{+}(v,\eta)$. The approximation in Eq~\eqref{eq:cap2collision_1} is based on a binomial expansion, which is justified by the fact that $p(v,\eta)$ is in general small. The second term accounts for the scenario in which exactly one collision occurs during the study. This occurs  with a probability of  $\frac{H}{\tau} \frac{L}{\eta v} p(v,\eta) \left(1-p(v,\eta)\right)^{\frac{H}{\tau} \frac{L}{\eta v} -1}$, and the corresponding throughput is reduced to $\left(1-\frac{T}{H}\right) s^{+}(v,\eta)$. A binomial expansion is also applied to the first approximation in Eq~\eqref{eq:cap2collision_2}, and the second approximation  holds under the assumption that $p(v,\eta)$ is sufficiently small such that higher-order terms beyond $p(v,\eta)^{2}$ can be negligible. 

The third term in Eq~\eqref{eq:cap2}, representing the two-accidents scenario, encloses two detailed situations. In the first situation, the two collisions  happens in the "Normal 1" or "Normal 2" regions of the original collision (see Figures~\ref{fig:overlapping} (a) and (b)) and thereby are overlapping. We denote the corresponding probability and the remaining capacity as $p_{co,2}$ and $s_{co,2}$, respectively. As previously analyzed, its remaining capacity is equivalent to that under the one-collision scenario, that is, $s_{co,2}=\left(1-\frac{T}{H}\right) s^{+}(v,\eta)$.

The second situation depicts two independent collisions, whose corresponding probability and the remaining capacity are represented as $p_{ci,2}$ and $s_{ci,2}$, respectively. As TCT is assumed to exceed half an hour, two independent collisions within the study period, which is set to one hour, would completely block the roadway, making the corresponding capacity $s_{ci,2}$ equal to $0$ veh/hr. Together, the third term in Eq~\eqref{eq:cap2} is given by:
\begin{flalign}
 p_{c,2}(v,\eta) s_{c,2}(v,\eta) &=p_{co,2}(v,\eta) s_{co,2}(v,\eta) + p_{ci,2}(v,\eta) s_{ci,2}(v,\eta) \notag \\
 &\approx \frac{\left(\frac{H}{\tau} \frac{L}{\eta v}\right)\left(\frac{H}{\tau} \frac{L}{\eta v}-1\right)}{2} p(v,\eta)^2 \left(1-\frac{T}{H}\right) s^{+}(v,\eta) \frac{1}{HL^2} \int_0^L \left( \frac{x^2}{2 v} + \frac{(L-x)^2}{2 c} \right) dx +0 \notag \\
 &= \frac{L(v+c)}{12 vcH} \left(\frac{H}{\tau} \frac{L}{\eta v}\right)\left(\frac{H}{\tau} \frac{L}{\eta v}-1\right) p(v,\eta)^2 \left(1-\frac{T}{H}\right) s^{+}(v,\eta). \label{eq:cap2collision_3}
\end{flalign} 

Similarly, the approximation is also based on binomial expansion with terms with equal or higher order than $p(v,\eta)^2$ since $p(v,\eta)$ is small.

Integrating Eqs~\eqref{eq:cap2collision_1}-\eqref{eq:cap2collision_3} together, the CIC considering overlapping collisions is summarized as follows:
\begin{flalign}
 s_c(v,\eta) \approx \left( 1 - \frac{T}{\tau} \frac{L}{\eta v} p(v,\eta) + \left( \frac{T}{H} - \frac{1}{2} + \frac{L(v+c)}{12 vcH} \left(1-\frac{T}{H}\right) \right) \left(\frac{H}{\tau} \frac{L}{\eta v}\right)\left(\frac{H}{\tau} \frac{L}{\eta v}-1\right) p(v,\eta)^2 \right) s^{+}(\eta). \label{eq:cap2collision}
\end{flalign}
\end{subequations}
\textbf{Remark}: Given that AV collisions are small probability events due to the stringent AV safety requirement, we assume that three collisions in one study period is nearly impossible and thereby neglecting the discussion of this scenario.

\subsection{Multiple-lane roads}
The previous analyses assume that all collisions occur within the same lane. On roads supporting fully autonomous traffic with multiple lanes, the CIC per lane remains unchanged when a single collision occurs. However, when two collisions occur, their relative positions across different lanes become critical in determining the macroscopic traffic performances. In this context, the property of the semi-Markov process diminishes, making it impossible to establish a stationary CIC. Instead, we rely on the lane-average expected throughput of the roadway segment to measure the mobility of multi-lane traffic. 
\begin{figure}[!ht]
 \centering\includegraphics[width=0.85\textwidth]{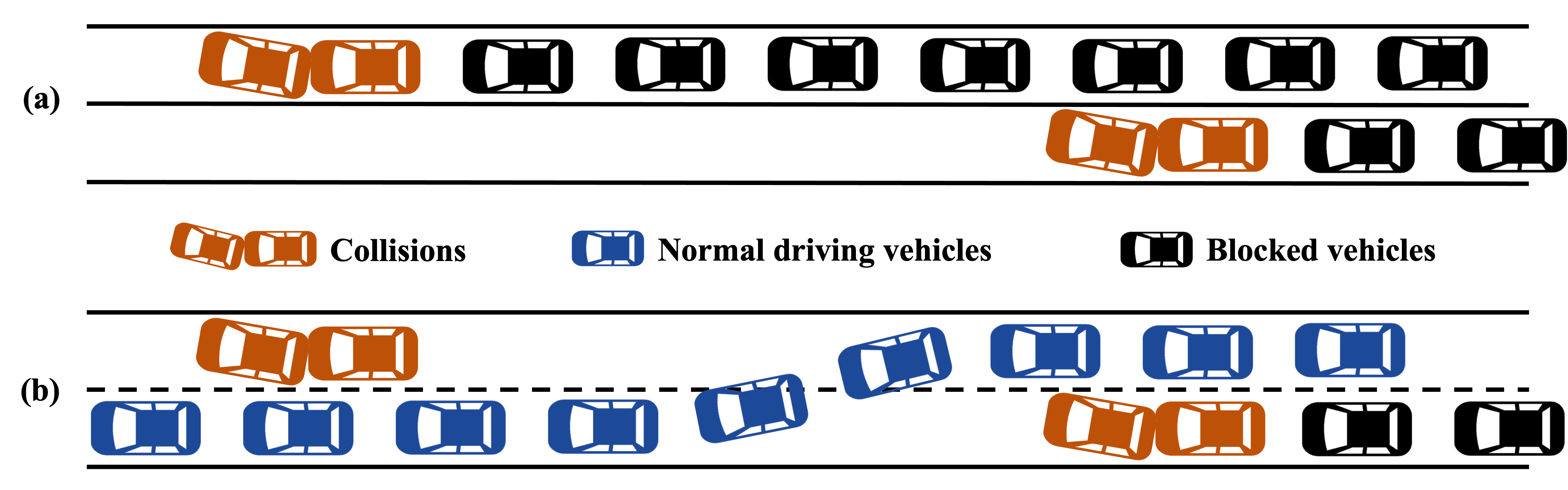}
 \caption{Two collisions occur on: (a) two independent lanes without lane-changing; (b) a two-lane road with lane-changing allowed.} \label{fig:dependent_2}
\end{figure}

In particular, we look into the scenario where two collisions occur on two adjacent lanes on a two-lane roadway. Figure~\ref{fig:dependent_2} compares the movements of AV strings under cases when AVs are allowed to change lanes or not. If lane changes are allowed, a throughput equals the full capacity of one lane can still be maintained. Accordingly, the lane-average expected throughput is given by:
\begin{subequations}
\begin{flalign}
 s_{l,2}(v, \eta) &= s_c(v,\eta) + p_{l,2}(v,\eta) \Delta s_{l,2}(v,\eta). \label{eq:cic_2lanes}
\end{flalign}

In this equation, $s_c$ stands for the expected throughput with not lane change permissions, which is identical to that in the single lane scenario given by Eq~\eqref{eq:cap2collision}. The second term represents the gains in throughput per lane from the lane change, where $p_{l,2}$ captures the probability when two collisions occurs on different lanes, and $\Delta s_{l,2}$ the expected additive capacity. Specifically,
\begin{flalign}
 p_{l,2}(v,\eta) &= \left( \frac{H}{\tau} \frac{L}{\eta v} p(v,\eta) \left(1-p(v,\eta)\right)^{\frac{H}{\tau} \frac{L}{\eta v}-1} \right)^2 \notag \\
  &\approx \frac{H^2}{\tau^2} \frac{L^2}{\eta^2 v^2} p(v,\eta)^2;
\end{flalign} \label{eq:cic_2lanes_p}
\begin{flalign}
 \Delta s_{l,2}(v,\eta) &= \frac{L-2D}{L} \frac{1}{H^2}\left(\int_0^{H-T}(T-t)dt + \int_{H-T}^{T}(2T-H)dt + \int_T^H(T-H+t)dt\right) \notag \\
 &\approx \frac{(2T-H+T)(H-T)+(T-H+T)(2T-H)}{2H^2} s^{+}(v,\eta) \notag \\
 &= \frac{T^2}{2H^2} s^{+}(\eta). \label{eq:cic_2lanes_s}
\end{flalign}

In Eq~\eqref{eq:cic_2lanes_s}, $D$ represents the minimum gap between two collisions that allow lane-changing, which is normally equal to several times the length of a vehicle. If there is not enough space to change lanes, the additive traffic throughput does not hold. However, compared with the road length $L$, $D$ is far small (i.e., $2D \ll L$). Hence we can ignore its marginal utility and derive the approximate expected capacity in the event of accidents in adjacent lanes.

Finally, the overall expected throughput per lane can then be derived by combining Eqs~\eqref{eq:cic_2lanes}-\eqref{eq:cic_2lanes_s}:
\begin{flalign}
 &s_{l,2}(v, \eta) \approx s_c(v,\eta) + \frac{1}{2}\frac{T^2}{\tau^2}\frac{L^2}{\eta^2v^2}p(v,\eta)^2 s^+(v,\eta) \notag \\
 &\approx \left( 1 - \frac{TL}{\tau\eta v} p(v,\eta) + \left( \left(HT-\frac{H^2}{2} + \frac{L(v+c)(H-T)}{12 vc} \right) \frac{L^2-L\tau\eta v}{\tau^2\eta^2 v^2} + \frac{T^2L^2}{2\tau^2\eta^2 v^2} \right) p(v,\eta)^2 \right) s^{+}(v,\eta). \label{eq:cic_2lanes_fin}
\end{flalign}
\end{subequations}

In conclusion, based on our above analyses, two-lane roads not only double the number of lanes but also slightly increase the expected throughput of each lane. Therefore, two-lane roads with lane change permissions have more than twice the traffic capacity compared to single-lane roads.

\subsection{Comparison}
Finally, we compare the hourly expected throughput in the extended scenarios with CIC derived under the baseline single-lane setting with non-overlapping collisions. Yet, these two metrics are not directly comparable, as the former represents an hourly expectation, while the latter reflects a long-term average under stationary conditions.To ensure a fair comparison, we first transform  CIC in Eq~\eqref{eq:s_e_fin} into its one-hour approximation by considering the case in which at most one collision occurs within an hour. This approximation is motivated by the consideration that two independent collisions would block the lane for the entire hour as each of their TCTs exceeds half an hour. Mathematically, the resulting expression corresponds to the first-order Taylor expansion of Eq~\eqref{eq:s_e_fin}. Consistent with the derivations shown in Eqs~\eqref{eq:cap2collision} and~\eqref{eq:cic_2lanes_fin}, it neglects higher-order terms, specifically those of order $p(v,\eta)^2$  and beyond. The final mathematically expression is given by:
\begin{flalign}
 s(v, \eta) &\approx \left( 1 - \frac{TL}{\tau\eta v} p(v,\eta) \right) s^{+}(v,\eta) + \frac{T^2 L^2}{\tau^2\eta^2 v^2} p(v,\eta)^2 s^{-}(v,\eta) \notag \\
 &= \left( 1 - \frac{TL}{\tau\eta v} p(v,\eta) \right) s^{+}(v,\eta). \label{eq:s_cic_approx}
\end{flalign}

\begin{figure}[htpb]
 \centering
 \includegraphics[width=0.65\textwidth]{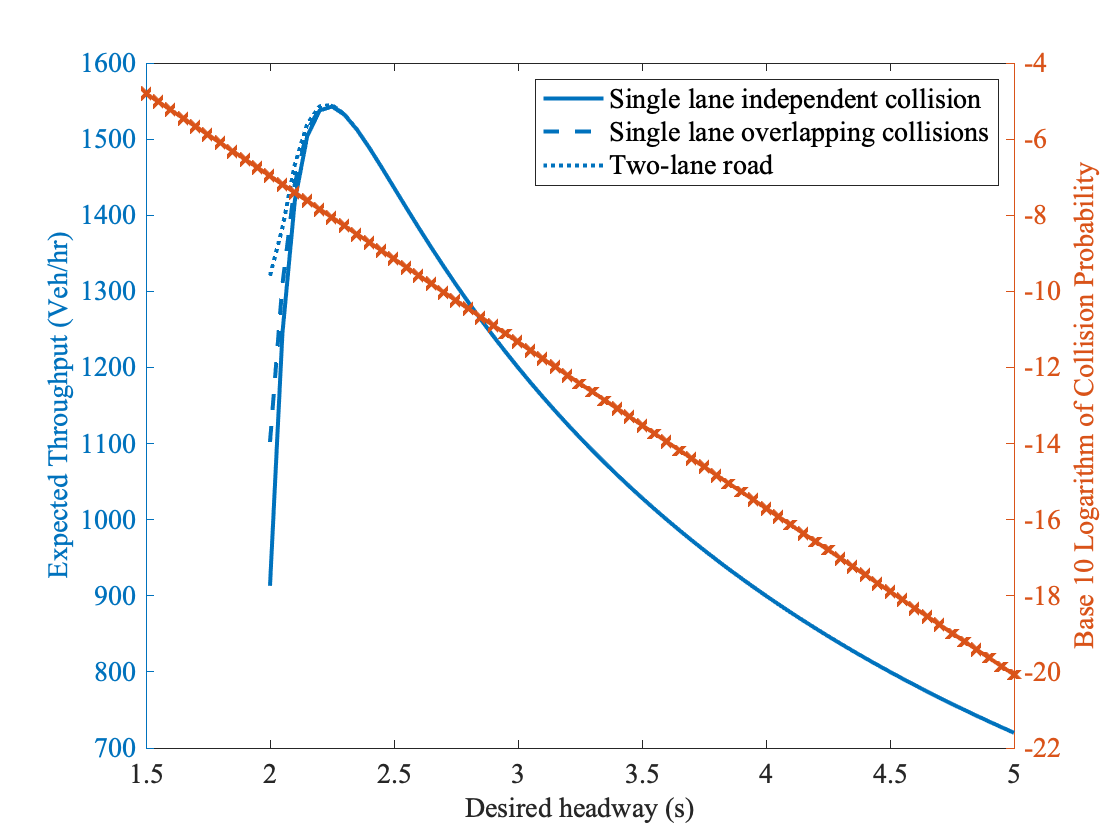}
 \caption{The comparison of the one-hour expected throughput (in blue curves) under three conditions with the same single-vehicle collision probability.}\label{fig:multi_comparison}
\end{figure}

Figure~\ref{fig:multi_comparison} presents the comparison results at an operational speed of 50 km/h. Compared to the baseline scenario, the presence of overlapping collisions and the allowance of lane changing on multi-lane roads both contribute to increased throughput when the desired headway is less than 2.5 seconds. Furthermore, multi-lane roads that permit lane changing yield a greater increase in throughput than overlapping collisions. When headway exceeds 2.5 seconds, the throughput gains in both scenarios become negligible. It is also noticed that the throughput under the extended scenarios retains the properties summarized in Section~\ref{sec:5}. In this regard, the proposed control and management strategies remains applicable to the extended scenarios. 

\section{Conclusion}\label{sec:7}
In this paper, we analyzed the impact of microscopic robotic uncertainties in autonomous vehicles on macroscopic traffic in terms of collisions and capacity in the car-following scenario. These systematic uncertainties rooted in sensors, onboard algorithms, and actuators of all mobile robots including AVs, which are unrelated to the external environment and driving intentions, contribute to their stochastic deviation from the intended movement trajectory. These random movements serve as a source of collisions, undermining the safety and mobility performance of the fully autonomous traffic system.

A modified IDM model is adopted to describe AVs' car-following behaviors under robotic errors, including perception and control errors, revealing that the resulting car-following distance follows a Gaussian distribution. This characteristic is further validated using real-world autonomous driving data from Waymo. The statistical nature also allows for an explicit formulation of the probability of rear-end collisions caused by uncertainties in the car-following headway. By incorporating the total clearance time required for a collision resolution, the dynamics of traffic capacity at each point are mathematically modeled as a function of speed and headway through a semi-Markov process. The expected value of capacity, referred to as collision-inclusive capacity, serves as a key metric for evaluating the combined mobility and safety performance of fully autonomous traffic. 

Further analyses investigate the influence of road length, vehicle dimensions, and the independent robotic uncertainties of AVs, offering valuable insights for AV development and adaptability to road networks. More importantly, two optimization problems are proposed for the macroscopic management. One determines the optimal headway and speed that maximizes collision-inclusive capacity under safety constraints, and the other minimizes the collision probability while ensuring collision-inclusive capacity meets demand. Theoretical discussions suggest that speed maximization under optimal conditions is recommended in both cases.

Our future work will extend the investigation into safety and mobility trade-offs in more AV-involved traffic operational scenarios, including those with complex traffic dynamics, as well as mixed traffic flows involving both AVs and HDVs, or heterogeneous AVs from different manufacturers. As robotic uncertainty is widely recognized in complex self-driving scenarios \citep{cao2023continuous},  our analytical model grounded in probabilistic characterizations is expected to remain applicable under these conditions.

In scenarios involving both AVs and HVs, it it worth noting that the randomness associated with human-driven vehicles has been investigated in both macroscopic traffic flow models \citep{jabari2012stochastic} and microscopic car-following models \citep{xu2020statistical}. Building on this existing body of research, our microscopic analysis can be extended  to incorporate the heterogeneous motion stochasticity for AVs and HDVs, contributing to a promising foundation for modeling compound uncertainty in mixed AV-HDV traffic.

Building on the richness of the proposed model framework, our future studies will also expand to discussions on the economic benefits, investment strategies, and managerial insights for AV development. Specifically, we will focus on optimizing key performance metrics, such as the maximum allowable collision probability, and the co-opetition between government agencies and AV manufacturers when their objectives differ.

\section*{Acknowledgment}
Dr. Xiaotong Sun and Chenglin Zhuang would like to thank National Science Foundation of China (72201073) and the Guangzhou Municipal Science and Technology Project (2023A03J0679, 2023A03J0011) for their partial supports to this project. 
Hangyu Li and Dr. Xiaopeng Li are not supported by these funding agencies.

\bibliographystyle{cas-model2-names}
\bibliography{cas-refs}

\begin{thebibliography}{63}
\expandafter\ifx\csname natexlab\endcsname\relax\def\natexlab#1{#1}\fi
\providecommand{\url}[1]{\texttt{#1}}
\providecommand{\href}[2]{#2}
\providecommand{\path}[1]{#1}
\providecommand{\DOIprefix}{doi:}
\providecommand{\ArXivprefix}{arXiv:}
\providecommand{\URLprefix}{URL: }
\providecommand{\Pubmedprefix}{pmid:}
\providecommand{\doi}[1]{\href{http://dx.doi.org/#1}{\path{#1}}}
\providecommand{\Pubmed}[1]{\href{pmid:#1}{\path{#1}}}
\providecommand{\bibinfo}[2]{#2}
\ifx\xfnm\relax \def\xfnm[#1]{\unskip,\space#1}\fi
\bibitem[{Araujo et~al.(2023)Araujo, Mousavi and Varshosaz}]{araujo2023testing}
\bibinfo{author}{Araujo, H.}, \bibinfo{author}{Mousavi, M.R.}, \bibinfo{author}{Varshosaz, M.}, \bibinfo{year}{2023}.
\newblock \bibinfo{title}{Testing, validation, and verification of robotic and autonomous systems: a systematic review}.
\newblock \bibinfo{journal}{ACM Transactions on Software Engineering and Methodology} \bibinfo{volume}{32}, \bibinfo{pages}{1--61}.
\bibitem[{Caesar et~al.(2019)Caesar, Bankiti, Lang, Vora, Liong, Xu, Krishnan, Pan, Baldan and Beijbom}]{nuscenes2019}
\bibinfo{author}{Caesar, H.}, \bibinfo{author}{Bankiti, V.}, \bibinfo{author}{Lang, A.H.}, \bibinfo{author}{Vora, S.}, \bibinfo{author}{Liong, V.E.}, \bibinfo{author}{Xu, Q.}, \bibinfo{author}{Krishnan, A.}, \bibinfo{author}{Pan, Y.}, \bibinfo{author}{Baldan, G.}, \bibinfo{author}{Beijbom, O.}, \bibinfo{year}{2019}.
\newblock \bibinfo{title}{nuscenes: A multimodal dataset for autonomous driving}.
\newblock \bibinfo{journal}{arXiv preprint arXiv:1903.11027} .
\bibitem[{Cao et~al.(2023)Cao, Jiang, Zhou, Xu, Peng and Yang}]{cao2023continuous}
\bibinfo{author}{Cao, Z.}, \bibinfo{author}{Jiang, K.}, \bibinfo{author}{Zhou, W.}, \bibinfo{author}{Xu, S.}, \bibinfo{author}{Peng, H.}, \bibinfo{author}{Yang, D.}, \bibinfo{year}{2023}.
\newblock \bibinfo{title}{Continuous improvement of self-driving cars using dynamic confidence-aware reinforcement learning}.
\newblock \bibinfo{journal}{Nature Machine Intelligence} \bibinfo{volume}{5}, \bibinfo{pages}{145--158}.
\bibitem[{Chen et~al.(2017)Chen, Ahn, Chitturi and Noyce}]{chen2017towards}
\bibinfo{author}{Chen, D.}, \bibinfo{author}{Ahn, S.}, \bibinfo{author}{Chitturi, M.}, \bibinfo{author}{Noyce, D.A.}, \bibinfo{year}{2017}.
\newblock \bibinfo{title}{Towards vehicle automation: Roadway capacity formulation for traffic mixed with regular and automated vehicles}.
\newblock \bibinfo{journal}{Transportation research part B: methodological} \bibinfo{volume}{100}, \bibinfo{pages}{196--221}.
\bibitem[{Christoforou et~al.(2010)Christoforou, Cohen and Karlaftis}]{christoforou2010vehicle}
\bibinfo{author}{Christoforou, Z.}, \bibinfo{author}{Cohen, S.}, \bibinfo{author}{Karlaftis, M.G.}, \bibinfo{year}{2010}.
\newblock \bibinfo{title}{Vehicle occupant injury severity on highways: An empirical investigation}.
\newblock \bibinfo{journal}{Accident Analysis \& Prevention} \bibinfo{volume}{42}, \bibinfo{pages}{1606--1620}.
\bibitem[{Daganzo(1997)}]{daganzo1997fundamentals}
\bibinfo{author}{Daganzo, C.F.}, \bibinfo{year}{1997}.
\newblock \bibinfo{title}{Fundamentals of transportation and traffic operations}.
\bibitem[{De~Gelder et~al.(2021)De~Gelder, Elrofai, Saberi, Paardekooper, Den~Camp and De~Schutter}]{de2021risk}
\bibinfo{author}{De~Gelder, E.}, \bibinfo{author}{Elrofai, H.}, \bibinfo{author}{Saberi, A.K.}, \bibinfo{author}{Paardekooper, J.P.}, \bibinfo{author}{Den~Camp, O.O.}, \bibinfo{author}{De~Schutter, B.}, \bibinfo{year}{2021}.
\newblock \bibinfo{title}{Risk quantification for automated driving systems in real-world driving scenarios}.
\newblock \bibinfo{journal}{IEEE Access} \bibinfo{volume}{9}, \bibinfo{pages}{168953--168970}.
\bibitem[{Dougald et~al.(2016)Dougald, Venkatanarayana, Goodall et~al.}]{dougald2016traffic}
\bibinfo{author}{Dougald, L.E.}, \bibinfo{author}{Venkatanarayana, R.}, \bibinfo{author}{Goodall, N.J.}, et~al., \bibinfo{year}{2016}.
\newblock \bibinfo{title}{Traffic incident management quick clearance guidance and implications}.
\newblock \bibinfo{type}{Technical Report}. Virginia Transportation Research Council.
\bibitem[{Ettinger et~al.(2021)Ettinger, Cheng, Caine, Liu, Zhao, Pradhan, Chai, Sapp, Qi, Zhou, Yang, Chouard, Sun, Ngiam, Vasudevan, McCauley, Shlens and Anguelov}]{Ettinger_2021_ICCV}
\bibinfo{author}{Ettinger, S.}, \bibinfo{author}{Cheng, S.}, \bibinfo{author}{Caine, B.}, \bibinfo{author}{Liu, C.}, \bibinfo{author}{Zhao, H.}, \bibinfo{author}{Pradhan, S.}, \bibinfo{author}{Chai, Y.}, \bibinfo{author}{Sapp, B.}, \bibinfo{author}{Qi, C.R.}, \bibinfo{author}{Zhou, Y.}, \bibinfo{author}{Yang, Z.}, \bibinfo{author}{Chouard, A.}, \bibinfo{author}{Sun, P.}, \bibinfo{author}{Ngiam, J.}, \bibinfo{author}{Vasudevan, V.}, \bibinfo{author}{McCauley, A.}, \bibinfo{author}{Shlens, J.}, \bibinfo{author}{Anguelov, D.}, \bibinfo{year}{2021}.
\newblock \bibinfo{title}{Large scale interactive motion forecasting for autonomous driving: The waymo open motion dataset}, in: \bibinfo{booktitle}{Proceedings of the IEEE/CVF International Conference on Computer Vision (ICCV)}, pp. \bibinfo{pages}{9710--9719}.
\bibitem[{Feng et~al.(2023)Feng, Sun, Yan, Zhu, Zou, Shen and Liu}]{feng2023dense}
\bibinfo{author}{Feng, S.}, \bibinfo{author}{Sun, H.}, \bibinfo{author}{Yan, X.}, \bibinfo{author}{Zhu, H.}, \bibinfo{author}{Zou, Z.}, \bibinfo{author}{Shen, S.}, \bibinfo{author}{Liu, H.X.}, \bibinfo{year}{2023}.
\newblock \bibinfo{title}{Dense reinforcement learning for safety validation of autonomous vehicles}.
\newblock \bibinfo{journal}{Nature} \bibinfo{volume}{615}, \bibinfo{pages}{620--627}.
\bibitem[{Feng et~al.(2021)Feng, Yan, Sun, Feng and Liu}]{feng2021intelligent}
\bibinfo{author}{Feng, S.}, \bibinfo{author}{Yan, X.}, \bibinfo{author}{Sun, H.}, \bibinfo{author}{Feng, Y.}, \bibinfo{author}{Liu, H.X.}, \bibinfo{year}{2021}.
\newblock \bibinfo{title}{Intelligent driving intelligence test for autonomous vehicles with naturalistic and adversarial environment}.
\newblock \bibinfo{journal}{Nature communications} \bibinfo{volume}{12}, \bibinfo{pages}{1--14}.
\bibitem[{Feng et~al.(2019)Feng, Zhang, Li, Cao, Liu and Li}]{feng2019string}
\bibinfo{author}{Feng, S.}, \bibinfo{author}{Zhang, Y.}, \bibinfo{author}{Li, S.E.}, \bibinfo{author}{Cao, Z.}, \bibinfo{author}{Liu, H.X.}, \bibinfo{author}{Li, L.}, \bibinfo{year}{2019}.
\newblock \bibinfo{title}{String stability for vehicular platoon control: Definitions and analysis methods}.
\newblock \bibinfo{journal}{Annual Reviews in Control} \bibinfo{volume}{47}, \bibinfo{pages}{81--97}.
\bibitem[{Fernandes and Nunes(2012)}]{fernandes2012platooning}
\bibinfo{author}{Fernandes, P.}, \bibinfo{author}{Nunes, U.}, \bibinfo{year}{2012}.
\newblock \bibinfo{title}{Platooning with ivc-enabled autonomous vehicles: Strategies to mitigate communication delays, improve safety and traffic flow}.
\newblock \bibinfo{journal}{IEEE Transactions on Intelligent Transportation Systems} \bibinfo{volume}{13}, \bibinfo{pages}{91--106}.
\bibitem[{Greenshields et~al.(1935)Greenshields, Bibbins, Channing and Miller}]{greenshields1935study}
\bibinfo{author}{Greenshields, B.D.}, \bibinfo{author}{Bibbins, J.}, \bibinfo{author}{Channing, W.}, \bibinfo{author}{Miller, H.}, \bibinfo{year}{1935}.
\newblock \bibinfo{title}{A study of traffic capacity}, in: \bibinfo{booktitle}{Highway research board proceedings}, \bibinfo{organization}{Washington, DC}. pp. \bibinfo{pages}{448--477}.
\bibitem[{Grigorev et~al.(2022)Grigorev, Mihaita, Lee and Chen}]{grigorev2022incident}
\bibinfo{author}{Grigorev, A.}, \bibinfo{author}{Mihaita, A.S.}, \bibinfo{author}{Lee, S.}, \bibinfo{author}{Chen, F.}, \bibinfo{year}{2022}.
\newblock \bibinfo{title}{Incident duration prediction using a bi-level machine learning framework with outlier removal and intra--extra joint optimisation}.
\newblock \bibinfo{journal}{Transportation Research Part C: Emerging Technologies} \bibinfo{volume}{141}, \bibinfo{pages}{103721}.
\bibitem[{Gunter et~al.(2020)Gunter, Gloudemans, Stern, McQuade, Bhadani, Bunting, Delle~Monache, Lysecky, Seibold, Sprinkle et~al.}]{gunter2020commercially}
\bibinfo{author}{Gunter, G.}, \bibinfo{author}{Gloudemans, D.}, \bibinfo{author}{Stern, R.E.}, \bibinfo{author}{McQuade, S.}, \bibinfo{author}{Bhadani, R.}, \bibinfo{author}{Bunting, M.}, \bibinfo{author}{Delle~Monache, M.L.}, \bibinfo{author}{Lysecky, R.}, \bibinfo{author}{Seibold, B.}, \bibinfo{author}{Sprinkle, J.}, et~al., \bibinfo{year}{2020}.
\newblock \bibinfo{title}{Are commercially implemented adaptive cruise control systems string stable?}
\newblock \bibinfo{journal}{IEEE Transactions on Intelligent Transportation Systems} \bibinfo{volume}{22}, \bibinfo{pages}{6992--7003}.
\bibitem[{Howard and Dai(2014)}]{howard2014public}
\bibinfo{author}{Howard, D.}, \bibinfo{author}{Dai, D.}, \bibinfo{year}{2014}.
\newblock \bibinfo{title}{Public perceptions of self-driving cars: The case of berkeley, california}, in: \bibinfo{booktitle}{Transportation research board 93rd annual meeting}, pp. \bibinfo{pages}{1--16}.
\bibitem[{Hu et~al.(2022)Hu, Xie, Xie, Lu, Xu and Su}]{hu2022distributed}
\bibinfo{author}{Hu, X.}, \bibinfo{author}{Xie, L.}, \bibinfo{author}{Xie, L.}, \bibinfo{author}{Lu, S.}, \bibinfo{author}{Xu, W.}, \bibinfo{author}{Su, H.}, \bibinfo{year}{2022}.
\newblock \bibinfo{title}{Distributed model predictive control for vehicle platoon with mixed disturbances and model uncertainties}.
\newblock \bibinfo{journal}{IEEE Transactions on Intelligent Transportation Systems} \bibinfo{volume}{23}, \bibinfo{pages}{17354--17365}.
\bibitem[{Huang et~al.(2024)Huang, Sheng, Ma and Chen}]{huang2024human}
\bibinfo{author}{Huang, Z.}, \bibinfo{author}{Sheng, Z.}, \bibinfo{author}{Ma, C.}, \bibinfo{author}{Chen, S.}, \bibinfo{year}{2024}.
\newblock \bibinfo{title}{Human as ai mentor: Enhanced human-in-the-loop reinforcement learning for safe and efficient autonomous driving}.
\newblock \bibinfo{journal}{Communications in Transportation Research} \bibinfo{volume}{4}, \bibinfo{pages}{100127}.
\bibitem[{Jabari and Liu(2012)}]{jabari2012stochastic}
\bibinfo{author}{Jabari, S.E.}, \bibinfo{author}{Liu, H.X.}, \bibinfo{year}{2012}.
\newblock \bibinfo{title}{A stochastic model of traffic flow: Theoretical foundations}.
\newblock \bibinfo{journal}{Transportation Research Part B: Methodological} \bibinfo{volume}{46}, \bibinfo{pages}{156--174}.
\bibitem[{Jim{\'e}nez et~al.(2016)Jim{\'e}nez, Naranjo, Anaya, Garc{\'\i}a, Ponz and Armingol}]{jimenez2016advanced}
\bibinfo{author}{Jim{\'e}nez, F.}, \bibinfo{author}{Naranjo, J.E.}, \bibinfo{author}{Anaya, J.J.}, \bibinfo{author}{Garc{\'\i}a, F.}, \bibinfo{author}{Ponz, A.}, \bibinfo{author}{Armingol, J.M.}, \bibinfo{year}{2016}.
\newblock \bibinfo{title}{Advanced driver assistance system for road environments to improve safety and efficiency}.
\newblock \bibinfo{journal}{Transportation research procedia} \bibinfo{volume}{14}, \bibinfo{pages}{2245--2254}.
\bibitem[{Knoop et~al.(2019)Knoop, Wang, Wilmink, Hoedemaeker, Maaskant and Van~der Meer}]{knoop2019platoon}
\bibinfo{author}{Knoop, V.L.}, \bibinfo{author}{Wang, M.}, \bibinfo{author}{Wilmink, I.}, \bibinfo{author}{Hoedemaeker, D.M.}, \bibinfo{author}{Maaskant, M.}, \bibinfo{author}{Van~der Meer, E.J.}, \bibinfo{year}{2019}.
\newblock \bibinfo{title}{Platoon of sae level-2 automated vehicles on public roads: Setup, traffic interactions, and stability}.
\newblock \bibinfo{journal}{Transportation Research Record} \bibinfo{volume}{2673}, \bibinfo{pages}{311--322}.
\bibitem[{Kontar and Ahn(2022)}]{kontar2022bayesian}
\bibinfo{author}{Kontar, W.}, \bibinfo{author}{Ahn, S.}, \bibinfo{year}{2022}.
\newblock \bibinfo{title}{Bayesian methods in automated vehicle's car-following uncertainties: Enabling strategic decision making}.
\newblock \bibinfo{journal}{arXiv preprint arXiv:2210.13683} .
\bibitem[{Kyriakidis et~al.(2015)Kyriakidis, Happee and de~Winter}]{kyriakidis2015public}
\bibinfo{author}{Kyriakidis, M.}, \bibinfo{author}{Happee, R.}, \bibinfo{author}{de~Winter, J.C.}, \bibinfo{year}{2015}.
\newblock \bibinfo{title}{Public opinion on automated driving: Results of an international questionnaire among 5000 respondents}.
\newblock \bibinfo{journal}{Transportation research part F: traffic psychology and behaviour} \bibinfo{volume}{32}, \bibinfo{pages}{127--140}.
\bibitem[{Lee et~al.(2024)Lee, Wang, Jang, Hayat, Bunting, Alanqary, Barbour, Fu, Gong, Gunter et~al.}]{lee2024traffic}
\bibinfo{author}{Lee, J.W.}, \bibinfo{author}{Wang, H.}, \bibinfo{author}{Jang, K.}, \bibinfo{author}{Hayat, A.}, \bibinfo{author}{Bunting, M.}, \bibinfo{author}{Alanqary, A.}, \bibinfo{author}{Barbour, W.}, \bibinfo{author}{Fu, Z.}, \bibinfo{author}{Gong, X.}, \bibinfo{author}{Gunter, G.}, et~al., \bibinfo{year}{2024}.
\newblock \bibinfo{title}{Traffic control via connected and automated vehicles: An open-road field experiment with 100 cavs}.
\newblock \bibinfo{journal}{arXiv preprint arXiv:2402.17043} .
\bibitem[{Levinson et~al.(2011)Levinson, Askeland, Becker, Dolson, Held, Kammel, Kolter, Langer, Pink, Pratt et~al.}]{levinson2011towards}
\bibinfo{author}{Levinson, J.}, \bibinfo{author}{Askeland, J.}, \bibinfo{author}{Becker, J.}, \bibinfo{author}{Dolson, J.}, \bibinfo{author}{Held, D.}, \bibinfo{author}{Kammel, S.}, \bibinfo{author}{Kolter, J.Z.}, \bibinfo{author}{Langer, D.}, \bibinfo{author}{Pink, O.}, \bibinfo{author}{Pratt, V.}, et~al., \bibinfo{year}{2011}.
\newblock \bibinfo{title}{Towards fully autonomous driving: Systems and algorithms}, in: \bibinfo{booktitle}{2011 IEEE intelligent vehicles symposium (IV)}, \bibinfo{organization}{IEEE}. pp. \bibinfo{pages}{163--168}.
\bibitem[{Li et~al.(2022)Li, Chen and Zhang}]{li2022equilibrium}
\bibinfo{author}{Li, J.}, \bibinfo{author}{Chen, D.}, \bibinfo{author}{Zhang, M.}, \bibinfo{year}{2022}.
\newblock \bibinfo{title}{Equilibrium modeling of mixed autonomy traffic flow based on game theory}.
\newblock \bibinfo{journal}{Transportation research part B: methodological} \bibinfo{volume}{166}, \bibinfo{pages}{110--127}.
\bibitem[{Li et~al.(2021)Li, Chen, Zhou, Laval and Xie}]{li2021car}
\bibinfo{author}{Li, T.}, \bibinfo{author}{Chen, D.}, \bibinfo{author}{Zhou, H.}, \bibinfo{author}{Laval, J.}, \bibinfo{author}{Xie, Y.}, \bibinfo{year}{2021}.
\newblock \bibinfo{title}{Car-following behavior characteristics of adaptive cruise control vehicles based on empirical experiments}.
\newblock \bibinfo{journal}{Transportation research part B: methodological} \bibinfo{volume}{147}, \bibinfo{pages}{67--91}.
\bibitem[{Li(2022)}]{li2022trade}
\bibinfo{author}{Li, X.}, \bibinfo{year}{2022}.
\newblock \bibinfo{title}{Trade-off between safety, mobility and stability in automated vehicle following control: An analytical method}.
\newblock \bibinfo{journal}{Transportation Research Part B: Methodological} \bibinfo{volume}{166}, \bibinfo{pages}{1--18}.
\bibitem[{Liu and Park(2021)}]{liu2021seeing}
\bibinfo{author}{Liu, J.}, \bibinfo{author}{Park, J.M.}, \bibinfo{year}{2021}.
\newblock \bibinfo{title}{“seeing is not always believing”: Detecting perception error attacks against autonomous vehicles}.
\newblock \bibinfo{journal}{IEEE Transactions on Dependable and Secure Computing} \bibinfo{volume}{18}, \bibinfo{pages}{2209--2223}.
\bibitem[{Ma et~al.(2022)Ma, Wang, Zuo, Hou, Li and Jiang}]{ma2022string}
\bibinfo{author}{Ma, K.}, \bibinfo{author}{Wang, H.}, \bibinfo{author}{Zuo, Z.}, \bibinfo{author}{Hou, Y.}, \bibinfo{author}{Li, X.}, \bibinfo{author}{Jiang, R.}, \bibinfo{year}{2022}.
\newblock \bibinfo{title}{String stability of automated vehicles based on experimental analysis of feedback delay and parasitic lag}.
\newblock \bibinfo{journal}{Transportation research part C: emerging technologies} \bibinfo{volume}{145}, \bibinfo{pages}{103927}.
\bibitem[{Makridis et~al.(2021)Makridis, Mattas, Anesiadou and Ciuffo}]{makridis2021openacc}
\bibinfo{author}{Makridis, M.}, \bibinfo{author}{Mattas, K.}, \bibinfo{author}{Anesiadou, A.}, \bibinfo{author}{Ciuffo, B.}, \bibinfo{year}{2021}.
\newblock \bibinfo{title}{Openacc. an open database of car-following experiments to study the properties of commercial acc systems}.
\newblock \bibinfo{journal}{Transportation research part C: emerging technologies} \bibinfo{volume}{125}, \bibinfo{pages}{103047}.
\bibitem[{McCarthy(2022)}]{mccarthy2022autonomous}
\bibinfo{author}{McCarthy, R.L.}, \bibinfo{year}{2022}.
\newblock \bibinfo{title}{Autonomous vehicle accident data analysis: California ol 316 reports: 2015--2020}.
\newblock \bibinfo{journal}{ASCE-ASME J Risk and Uncert in Engrg Sys Part B Mech Engrg} \bibinfo{volume}{8}.
\bibitem[{Meng et~al.(2024)Meng, Li, Ornik and Li}]{meng2024koopman}
\bibinfo{author}{Meng, Y.}, \bibinfo{author}{Li, H.}, \bibinfo{author}{Ornik, M.}, \bibinfo{author}{Li, X.}, \bibinfo{year}{2024}.
\newblock \bibinfo{title}{Koopman-based data-driven techniques for adaptive cruise control system identification}, in: \bibinfo{booktitle}{27th IEEE International Conference on Intelligent Transportation Systems (ITSC), IEEE}.
\bibitem[{Mihaita et~al.(2019)Mihaita, Liu, Cai and Rizoiu}]{mihaita2019arterial}
\bibinfo{author}{Mihaita, A.S.}, \bibinfo{author}{Liu, Z.}, \bibinfo{author}{Cai, C.}, \bibinfo{author}{Rizoiu, M.A.}, \bibinfo{year}{2019}.
\newblock \bibinfo{title}{Arterial incident duration prediction using a bi-level framework of extreme gradient-tree boosting}.
\newblock \bibinfo{journal}{arXiv preprint arXiv:1905.12254} .
\bibitem[{Milan{\'e}s and Shladover(2014)}]{milanes2014modeling}
\bibinfo{author}{Milan{\'e}s, V.}, \bibinfo{author}{Shladover, S.E.}, \bibinfo{year}{2014}.
\newblock \bibinfo{title}{Modeling cooperative and autonomous adaptive cruise control dynamic responses using experimental data}.
\newblock \bibinfo{journal}{Transportation Research Part C: Emerging Technologies} \bibinfo{volume}{48}, \bibinfo{pages}{285--300}.
\bibitem[{Mohammadian et~al.(2023)Mohammadian, Zheng, Haque and Bhaskar}]{mohammadian2023continuum}
\bibinfo{author}{Mohammadian, S.}, \bibinfo{author}{Zheng, Z.}, \bibinfo{author}{Haque, M.M.}, \bibinfo{author}{Bhaskar, A.}, \bibinfo{year}{2023}.
\newblock \bibinfo{title}{Continuum modeling of freeway traffic flows: State-of-the-art, challenges and future directions in the era of connected and automated vehicles}.
\newblock \bibinfo{journal}{Communications in Transportation Research} \bibinfo{volume}{3}, \bibinfo{pages}{100107}.
\bibitem[{Morando et~al.(2018)Morando, Tian, Truong and Vu}]{morando2018studying}
\bibinfo{author}{Morando, M.M.}, \bibinfo{author}{Tian, Q.}, \bibinfo{author}{Truong, L.T.}, \bibinfo{author}{Vu, H.L.}, \bibinfo{year}{2018}.
\newblock \bibinfo{title}{Studying the safety impact of autonomous vehicles using simulation-based surrogate safety measures}.
\newblock \bibinfo{journal}{Journal of advanced transportation} \bibinfo{volume}{2018}.
\bibitem[{Motors(2018)}]{general2018self}
\bibinfo{author}{Motors, G.}, \bibinfo{year}{2018}.
\newblock \bibinfo{title}{2018 self-driving safety report}.
\newblock \URLprefix \url{https://www.gm.com/content/dam/company/docs/us/en/gmcom/gmsafetyreport.pdf}.
\bibitem[{Mueller et~al.(2020)Mueller, Cicchino and Zuby}]{mueller2020humanlike}
\bibinfo{author}{Mueller, A.S.}, \bibinfo{author}{Cicchino, J.B.}, \bibinfo{author}{Zuby, D.S.}, \bibinfo{year}{2020}.
\newblock \bibinfo{title}{What humanlike errors do autonomous vehicles need to avoid to maximize safety?}
\newblock \bibinfo{journal}{Journal of safety research} \bibinfo{volume}{75}, \bibinfo{pages}{310--318}.
\bibitem[{Ni et~al.(2009)Ni, Ramanathan, Chehade, Balzano, Nair, Zahedi, Kohler, Pottie, Hansen and Srivastava}]{ni2009sensor}
\bibinfo{author}{Ni, K.}, \bibinfo{author}{Ramanathan, N.}, \bibinfo{author}{Chehade, M.N.H.}, \bibinfo{author}{Balzano, L.}, \bibinfo{author}{Nair, S.}, \bibinfo{author}{Zahedi, S.}, \bibinfo{author}{Kohler, E.}, \bibinfo{author}{Pottie, G.}, \bibinfo{author}{Hansen, M.}, \bibinfo{author}{Srivastava, M.}, \bibinfo{year}{2009}.
\newblock \bibinfo{title}{Sensor network data fault types}.
\newblock \bibinfo{journal}{ACM Transactions on Sensor Networks (TOSN)} \bibinfo{volume}{5}, \bibinfo{pages}{1--29}.
\bibitem[{Qin et~al.(2023)Qin, Luo and Wang}]{qin2023stability}
\bibinfo{author}{Qin, Y.}, \bibinfo{author}{Luo, Q.}, \bibinfo{author}{Wang, H.}, \bibinfo{year}{2023}.
\newblock \bibinfo{title}{Stability analysis and connected vehicles management for mixed traffic flow with platoons of connected automated vehicles}.
\newblock \bibinfo{journal}{Transportation Research Part C: Emerging Technologies} \bibinfo{volume}{157}, \bibinfo{pages}{104370}.
\bibitem[{Ran and Tsao(1996)}]{ran1996traffic}
\bibinfo{author}{Ran, B.}, \bibinfo{author}{Tsao, H.S.J.}, \bibinfo{year}{1996}.
\newblock \bibinfo{title}{Traffic flow analysis for an automated highway system} .
\bibitem[{Schakel et~al.(2017)Schakel, Gorter, De~Winter and Van~Arem}]{schakel2017driving}
\bibinfo{author}{Schakel, W.J.}, \bibinfo{author}{Gorter, C.M.}, \bibinfo{author}{De~Winter, J.C.}, \bibinfo{author}{Van~Arem, B.}, \bibinfo{year}{2017}.
\newblock \bibinfo{title}{Driving characteristics and adaptive cruise control? a naturalistic driving study}.
\newblock \bibinfo{journal}{IEEE Intelligent Transportation Systems Magazine} \bibinfo{volume}{9}, \bibinfo{pages}{17--24}.
\bibitem[{Seo and Asakura(2017)}]{seo2017endogenous}
\bibinfo{author}{Seo, T.}, \bibinfo{author}{Asakura, Y.}, \bibinfo{year}{2017}.
\newblock \bibinfo{title}{Endogenous market penetration dynamics of automated and connected vehicles: Transport-oriented model and its paradox}.
\newblock \bibinfo{journal}{Transportation Research Procedia} \bibinfo{volume}{27}, \bibinfo{pages}{238--245}.
\bibitem[{Shi and Li(2021)}]{shi2021empirical}
\bibinfo{author}{Shi, X.}, \bibinfo{author}{Li, X.}, \bibinfo{year}{2021}.
\newblock \bibinfo{title}{Empirical study on car-following characteristics of commercial automated vehicles with different headway settings}.
\newblock \bibinfo{journal}{Transportation research part C: emerging technologies} \bibinfo{volume}{128}, \bibinfo{pages}{103134}.
\bibitem[{Shladover and Nowakowski(2019)}]{shladover2019regulatory}
\bibinfo{author}{Shladover, S.E.}, \bibinfo{author}{Nowakowski, C.}, \bibinfo{year}{2019}.
\newblock \bibinfo{title}{Regulatory challenges for road vehicle automation: Lessons from the california experience}.
\newblock \bibinfo{journal}{Transportation research part A: policy and practice} \bibinfo{volume}{122}, \bibinfo{pages}{125--133}.
\bibitem[{Smith and Smith(2002)}]{smith2002forecasting}
\bibinfo{author}{Smith, K.}, \bibinfo{author}{Smith, B.L.}, \bibinfo{year}{2002}.
\newblock \bibinfo{title}{Forecasting the clearance time of freeway accidents}.
\newblock \bibinfo{type}{Technical Report} \bibinfo{number}{STL-2001-01}. Center for Transportation Studies, University of Virginia.
\bibitem[{Sybis et~al.(2020)Sybis, Rodziewicz and Weso{\l}owski}]{sybis2020influence}
\bibinfo{author}{Sybis, M.}, \bibinfo{author}{Rodziewicz, M.}, \bibinfo{author}{Weso{\l}owski, K.}, \bibinfo{year}{2020}.
\newblock \bibinfo{title}{Influence of sensor inaccuracies and acceleration limits on ieee 802.11 p-based cacc controlled platoons}, in: \bibinfo{booktitle}{2020 IEEE 91st Vehicular Technology Conference (VTC2020-Spring)}, \bibinfo{organization}{IEEE}. pp. \bibinfo{pages}{1--6}.
\bibitem[{Talebpour and Mahmassani(2016)}]{talebpour2016influence}
\bibinfo{author}{Talebpour, A.}, \bibinfo{author}{Mahmassani, H.S.}, \bibinfo{year}{2016}.
\newblock \bibinfo{title}{Influence of connected and autonomous vehicles on traffic flow stability and throughput}.
\newblock \bibinfo{journal}{Transportation Research Part C: Emerging Technologies} \bibinfo{volume}{71}, \bibinfo{pages}{143--163}.
\bibitem[{Treiber and Kesting(2013)}]{treiber2013traffic}
\bibinfo{author}{Treiber, M.}, \bibinfo{author}{Kesting, A.}, \bibinfo{year}{2013}.
\newblock \bibinfo{title}{Traffic flow dynamics}.
\newblock \bibinfo{journal}{Traffic Flow Dynamics: Data, Models and Simulation, Springer-Verlag Berlin Heidelberg} , \bibinfo{pages}{983--1000}.
\bibitem[{Vignon and Bahrami(2025)}]{vignon2025safety}
\bibinfo{author}{Vignon, D.}, \bibinfo{author}{Bahrami, S.}, \bibinfo{year}{2025}.
\newblock \bibinfo{title}{Safety, liability, and insurance markets in the age of automated driving}.
\newblock \bibinfo{journal}{Transportation Research Part B: Methodological} \bibinfo{volume}{191}, \bibinfo{pages}{103115}.
\bibitem[{Wang et~al.(2020)Wang, Li, Tian and Jiang}]{wang2020stability}
\bibinfo{author}{Wang, Y.}, \bibinfo{author}{Li, X.}, \bibinfo{author}{Tian, J.}, \bibinfo{author}{Jiang, R.}, \bibinfo{year}{2020}.
\newblock \bibinfo{title}{Stability analysis of stochastic linear car-following models}.
\newblock \bibinfo{journal}{Transportation Science} \bibinfo{volume}{54}, \bibinfo{pages}{274--297}.
\bibitem[{Waymo(2017)}]{waymo2017waymo}
\bibinfo{author}{Waymo}, \bibinfo{year}{2017}.
\newblock \bibinfo{title}{Waymo safety report: On the road to fully self-driving}.
\newblock \URLprefix \url{https://storage.googleapis.com/sdc-prod/v1/safety-report/waymo-safety-report-2017.pdf}.
\bibitem[{Wen et~al.(2022)Wen, Jiang, Wijaya, Li, Yang and Yang}]{wen2022tm3loc}
\bibinfo{author}{Wen, T.}, \bibinfo{author}{Jiang, K.}, \bibinfo{author}{Wijaya, B.}, \bibinfo{author}{Li, H.}, \bibinfo{author}{Yang, M.}, \bibinfo{author}{Yang, D.}, \bibinfo{year}{2022}.
\newblock \bibinfo{title}{Tm$^3$loc: Tightly-coupled monocular map matching for high precision vehicle localization}.
\newblock \bibinfo{journal}{IEEE Transactions on Intelligent Transportation Systems} .
\bibitem[{Xing et~al.(2022)Xing, Han, Zhang, Lu and Gao}]{xing2022bicyclists}
\bibinfo{author}{Xing, Y.}, \bibinfo{author}{Han, X.}, \bibinfo{author}{Zhang, H.M.}, \bibinfo{author}{Lu, J.}, \bibinfo{author}{Gao, Z.Y.}, \bibinfo{year}{2022}.
\newblock \bibinfo{title}{Do bicyclists and pedestrians support their city as an autonomous vehicle proving ground? evidence from pittsburgh}.
\newblock \bibinfo{journal}{Case Studies on Transport Policy} \bibinfo{volume}{10}, \bibinfo{pages}{2401--2412}.
\bibitem[{Xu et~al.(2022)Xu, Ding, Lyu, Liu, Wang, He, Hu, Zhao and Li}]{xu2023safebench}
\bibinfo{author}{Xu, C.}, \bibinfo{author}{Ding, W.}, \bibinfo{author}{Lyu, W.}, \bibinfo{author}{Liu, Z.}, \bibinfo{author}{Wang, S.}, \bibinfo{author}{He, Y.}, \bibinfo{author}{Hu, H.}, \bibinfo{author}{Zhao, D.}, \bibinfo{author}{Li, B.}, \bibinfo{year}{2022}.
\newblock \bibinfo{title}{Safebench: A benchmarking platform for safety evaluation of autonomous vehicles}, in: \bibinfo{booktitle}{Thirty-sixth Conference on Neural Information Processing Systems Datasets and Benchmarks Track}, pp. \bibinfo{pages}{25667--25682}.
\bibitem[{Xu and Laval(2020)}]{xu2020statistical}
\bibinfo{author}{Xu, T.}, \bibinfo{author}{Laval, J.}, \bibinfo{year}{2020}.
\newblock \bibinfo{title}{Statistical inference for two-regime stochastic car-following models}.
\newblock \bibinfo{journal}{Transportation Research Part B: Methodological} \bibinfo{volume}{134}, \bibinfo{pages}{210--228}.
\bibitem[{Yan et~al.(2023)Yan, Zou, Feng, Zhu, Sun and Liu}]{yan2023learning}
\bibinfo{author}{Yan, X.}, \bibinfo{author}{Zou, Z.}, \bibinfo{author}{Feng, S.}, \bibinfo{author}{Zhu, H.}, \bibinfo{author}{Sun, H.}, \bibinfo{author}{Liu, H.X.}, \bibinfo{year}{2023}.
\newblock \bibinfo{title}{Learning naturalistic driving environment with statistical realism}.
\newblock \bibinfo{journal}{Nature Communications} \bibinfo{volume}{14}, \bibinfo{pages}{2037}.
\bibitem[{Zhang et~al.(2020)Zhang, Lu, Yu, Wang and Yang}]{zhang2020effect}
\bibinfo{author}{Zhang, J.}, \bibinfo{author}{Lu, G.}, \bibinfo{author}{Yu, H.}, \bibinfo{author}{Wang, Y.}, \bibinfo{author}{Yang, C.}, \bibinfo{year}{2020}.
\newblock \bibinfo{title}{Effect of the uncertainty level of vehicle-position information on the stability and safety of the car-following process}.
\newblock \bibinfo{journal}{IEEE Transactions on Intelligent Transportation Systems} \bibinfo{volume}{23}, \bibinfo{pages}{4944--4958}.
\bibitem[{Zhou et~al.(2024)Zhou, Ma, Liang, Li and Qu}]{zhou2024unified}
\bibinfo{author}{Zhou, H.}, \bibinfo{author}{Ma, K.}, \bibinfo{author}{Liang, S.}, \bibinfo{author}{Li, X.}, \bibinfo{author}{Qu, X.}, \bibinfo{year}{2024}.
\newblock \bibinfo{title}{A unified longitudinal trajectory dataset for automated vehicle}.
\newblock \bibinfo{journal}{Scientific Data} \bibinfo{volume}{11}, \bibinfo{pages}{1123}.
\bibitem[{Zhou and Ahn(2019)}]{zhou2019robust}
\bibinfo{author}{Zhou, Y.}, \bibinfo{author}{Ahn, S.}, \bibinfo{year}{2019}.
\newblock \bibinfo{title}{Robust local and string stability for a decentralized car following control strategy for connected automated vehicles}.
\newblock \bibinfo{journal}{Transportation Research Part B: Methodological} \bibinfo{volume}{125}, \bibinfo{pages}{175--196}.
\bibitem[{Zhou et~al.(2017)Zhou, Ahn, Chitturi and Noyce}]{zhou2017rolling}
\bibinfo{author}{Zhou, Y.}, \bibinfo{author}{Ahn, S.}, \bibinfo{author}{Chitturi, M.}, \bibinfo{author}{Noyce, D.A.}, \bibinfo{year}{2017}.
\newblock \bibinfo{title}{Rolling horizon stochastic optimal control strategy for acc and cacc under uncertainty}.
\newblock \bibinfo{journal}{Transportation Research Part C: Emerging Technologies} \bibinfo{volume}{83}, \bibinfo{pages}{61--76}.

\end{thebibliography}

\end{document}